\documentclass[preprint,12pt,authoryear]{elsarticle}

\usepackage{amssymb}
\usepackage{amsmath}

\usepackage{epstopdf, epsfig}
\usepackage{amstext}
\usepackage{amsfonts,amssymb}
\usepackage{algorithm}
\usepackage{algorithmicx}
\usepackage{bm}
\usepackage{graphicx}
\usepackage{subcaption}
\usepackage{color}
\usepackage{subcaption}
\usepackage[HTML]{xcolor}

\definecolor{steelcyan}{RGB}{70,130,180}
\definecolor{mygray}{RGB}{89,89,89}
\definecolor{mypink}{RGB}{255,191,204}
\definecolor{mycyan}{RGB}{51,179,230}
\definecolor{EnkfColor}{HTML}{1f77b4}
\definecolor{EnKFMLCcolor}{HTML}{ff7f0e}

\usepackage{booktabs} 

\DeclareMathAlphabet{\pazocal}{OMS}{zplm}{m}{n}

    \def\dashL{\bm{\mbox{--~--~--}}}

\journal{Computer Methods in Applied Mechanics and Engineering}

\begin{document}

\begin{frontmatter}



\title{Small Ensemble-based Data Assimilation: A Machine Learning-Enhanced Data Assimilation Method with Limited Ensemble Size} 


\author[1]{Z. Li} 
\author[1]{Z. Yao}
\author[1]{X. Li}
\author[2]{Z. Liu}
\author[3]{Z. Lu}
\author[4]{S. X}
\author[5]{S. Kim}
\author[1,6,7]{G. Wang\corref{cor1}}

\affiliation[1]{organization={Centre for Regional Oceans, Department of Ocean Science and Technology, and State Key Laboratory of Internet of Things for Smart City, University of Macau},
            city={Macau}}
\affiliation[2]{organization={School of Naval Architecture and Ocean Engineering, Huazhong University of Science and Technology},
            city={Wuhan},
            state={Hubei},
            country={China}}
\affiliation[3]{organization={Ningbo Institute of Dalian University of Technology},
            city={Ningbo},
            state={Zhejiang},
            country={China}}
\affiliation[4]{organization={College of Civil Engineering, Zhejiang University of Technology},
            city={Hangzhou},
            state={Zhejiang},
            country={China}}
\affiliation[5]{organization={Department of Naval Architecture and Ocean Engineering, Hongik University},
            city={Sejong},
            country={Republic of Korea}}
\affiliation[6]{organization={State Key Laboratory of Internet of Things for Smart City, University of Macau},
            city={Macau}}
\affiliation[7]{organization={Zhuhai UM Science and Technology Research Institute},
            city={Zhuhai},
            state={Guangdong},
            country={China}}

\cortext[cor1]{Corresponding author.}
\ead{wanggy@um.edu.mo}

\begin{abstract}
Ensemble-based data assimilation (DA) methods have become increasingly popular due to their inherent ability to address nonlinear dynamic problems. However, these methods often face a trade-off between analysis accuracy and computational efficiency, as larger ensemble sizes required for higher accuracy also lead to greater computational cost. In this study, we propose a novel machine learning-based data assimilation approach that combines the traditional ensemble Kalman filter (EnKF) with a fully connected neural network (FCNN). Specifically, our method uses a relatively small ensemble size to generate preliminary yet suboptimal analysis states via EnKF. A FCNN is then employed to learn and predict correction terms for these states, thereby mitigating the performance degradation induced by the limited ensemble size. We evaluate the performance of our proposed EnKF–FCNN method through numerical experiments involving Lorenz systems and nonlinear ocean wave field simulations. The results consistently demonstrate that the new method achieves higher accuracy than traditional EnKF with the same ensemble size, while incurring negligible additional computational cost. Moreover, the EnKF-FCNN method is adaptable to diverse applications through coupling with different models and the use of alternative ensemble-based DA methods.
\end{abstract}







\end{frontmatter}



\section{Introduction}
\label{sec:Intro}
By solving a Bayesian estimation problem, data assimilation (DA) is a mathematical technique that combines model predictions and observations to obtain the optimal state estimate of dynamic systems~\citep{carrassi2018data}. Also, with the advancement of computation and observation technologies in the past few decades, DA methods have been commonly applied in various fields, such as meteorology~\citep{bocquet2015data,eyre2022assimilation}, physical oceanography~\citep{wang2021phase,wang2022phase,martin2025data}, land surface processes~\citep{li2024land,felsberg2021global}, and so on. With the inherent ability to describe the evolution of state error covariance, ensemble-based methods have become increasingly popular in addressing nonlinear problems, among which the ensemble Kalman filter (EnKF) is probably the most famous one with the non-intrusive nature and sequential DA capability~\citep{evensen1994sequential,evensen2003ensemble}.

Despite their wide applications, ensemble-based DA methods are still facing one intractable problem, which is the trade-off between the analysis accuracy and computational efficiency~\citep{petrie2010ensemble,whitaker2012evaluating}. Specifically, a relatively large ensemble provides a better representation of the uncertainty, particularly the covariance matrices, leading to more accurate state estimations and vice versa. But it should be noted that the computational cost scales (almost) linearly with the ensemble size. Consequently, the computational burden can become unsustainable or impractical for high-dimensional nonlinear systems. Therefore, it is an inevitable problem to balance the desire for high accuracy with the practical limitations of available computational resources. From a statistical perspective, a limited ensemble size can lead to a rank-deficiency issue of the estimated covariance matrices, often resulting in spurious correlations and underestimation and eventually causing divergence. In this regard, two ad hoc methods have been developed to address the rank deficiency problem, i.e. localization~\citep{buehner2007spectral} and inflation~\citep{kang2012estimation}. While these methods, along with their advanced variants, e.g. adaptive localization and inflation~\citep{evensen2022localization}, can mitigate performance degradation induced by a limited ensemble size, substantial efforts are still required to optimize their configurations~\citep{choi2025sampling}, which would inevitably increase the overall computational complexity.

The purpose of this paper is to show that the dilemma of ensemble-based DA methods can be significantly mitigated by machine learning (ML) methods, particularly neural networks (NNs). NNs are characterized by their ability to automatically learn and extract useful information from datasets, employing the composition of linear transformations with nonlinear activation functions to model complex relationships between inputs and outputs. Currently, NNs have shown potential to contribute to both the forecast and analysis steps involved in the DA-based simulations, although related research remains limited. In the forecast step, NNs are primarily used to build surrogate models that approximate the traditional, computationally expensive physical models, thereby accelerating the computational process~\citep{brajard2020combining,sun2025online}. However, this approach can suffer from reduced accuracy and limited generalization ability when encountering new or extreme scenarios not well represented in the training dataset. Regarding the analysis step, NNs are mainly applied to learn the update process (e.g. EnKF formula) for handling missing or low-quality data~\citep{arcucci2021deep,wu2021fast}. However, these methods rely on accurate estimates of covariance matrices, which typically depend on either a relatively large ensemble, ad hoc operations, or both.

In this study, we propose a novel small ensemble-based DA framework with machine learning, which can overcome the systematic biases induced by a limited ensemble size effectively. Specifically, we develop the machine learning-enhanced DA framework on the basis of EnKF, initialized with a limited ensemble size. Then, a fully connected neural network (FCNN) is employed to learn a correction term for the analysis state, which is aimed at mitigating the suboptimal state estimation caused by the limited ensemble size. Consequently, the EnKF-FCNN coupled algorithm can offer both favorable computational efficiency and accuracy simultaneously. To evaluate the performance of the EnKF–FCNN coupled algorithm, we perform a set of numerical experiments based on the simulations of Lorenz systems (Lorenz-63 and Lorenz-96) and nonlinear ocean wave fields, which consistently demonstrate higher accuracy with negligible additional computational cost compared to traditional EnKF when using the same limited ensemble size. Moreover, the EnKF-FCNN method is highly adaptable across various applications by integrating with different models and employing alternative ensemble-based DA methods.

The paper is organized as follows. $\S$~\ref{sec:CoupledAlgorithms} introduces the traditional EnKF and EnKF-FCNN coupled algorithms. The validation and benchmark of the EnKF-FCNN coupled algorithm based on the simulations of Lorenz systems and nonlinear ocean wave fields are presented in $\S$~\ref{sec:Experiment results}. Finally, $\S$~\ref{sec:Conclusion} briefly summarizes this work.

\section{Methods}
\label{sec:CoupledAlgorithms}

\subsection{Ensemble Kalman filter}
\label{sec:EnKF}
\begin{figure}[h]
    \centering
\includegraphics[width=0.9\textwidth,  trim=8cm 0cm 8cm 0cm, clip]{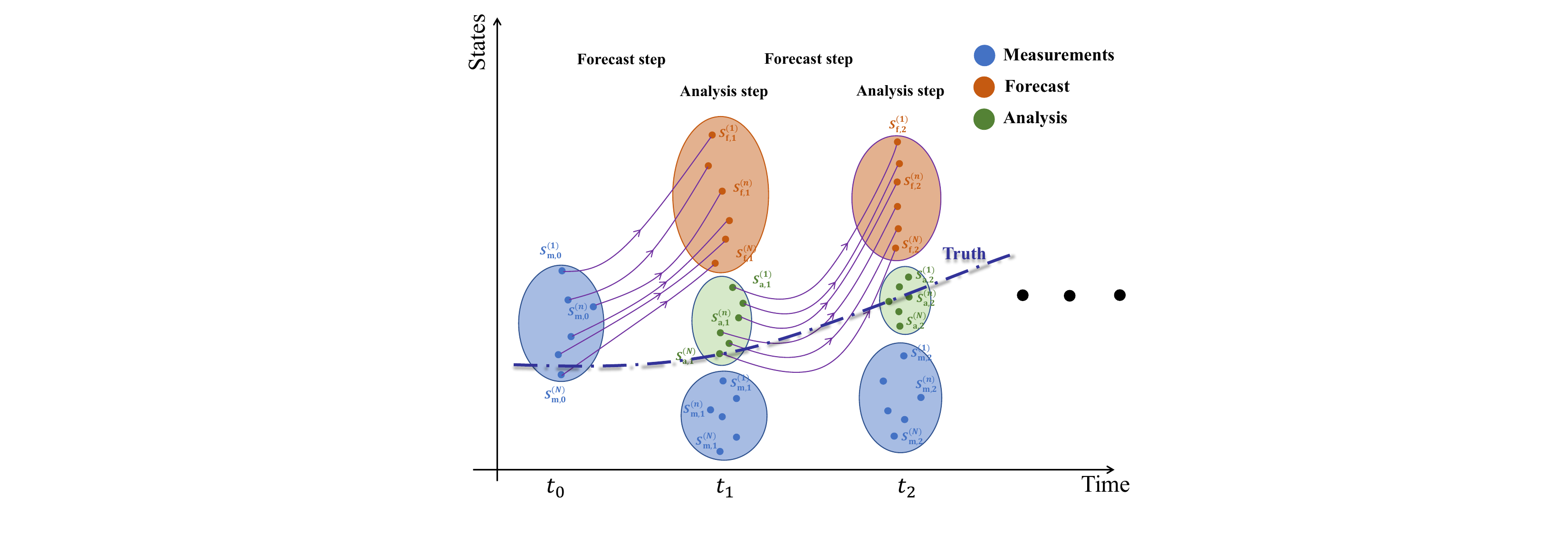}
    \caption{Schematic illustration of EnKF (reproduced with modifications from~\cite{wang2021phase}). The size of ellipses represents the
level of uncertainty. We use the notation $\boldsymbol{s}^{(n)}_{*,j}$ to represent the $n_{th}$ ($n=1,2,\dots N$) ensemble member state at time $t=t_j$, $j=0,1,2\dots$, with $*=m,f,a$ for measurement, forecast, and analysis.}
    \label{fig:ENKF}
\end{figure}

In this study, we use the stochastic EnKF~\citep{burgers1998analysis,van2020consistent} as the basis to develop the small ensemble-based DA method with machine learning, which basically includes the forecast and analysis steps as shown in Fig.~\ref{fig:ENKF}. At the beginning time of the simulation $t=t_0$, an ensemble of initial conditions is first generated, usually based on the prescribed distribution information of the initial condition~\citep{wang2021phase},
\begin{equation}
\boldsymbol{S}_{0}=\left[\boldsymbol{s}^{(1)}_{ 0},\boldsymbol{s}^{(2)}_{0},\dots \boldsymbol{s}^{(n)}_{0}, \dots \boldsymbol{s}^{(N-1)}_{0},\boldsymbol{s}^{(N)}_{0}\right],
\end{equation}
where $\boldsymbol{s}$ represents the state vector and $\boldsymbol{S}$ is the corresponding ensemble with size $N$ and index $n=1, 2, \dots N$. Then with the forward model from time $t_j$ to $t_{j+1}$, $\mathcal{M}_{j+1:j}$, the forecast step is performed by conducting $N$ simulations (here $j=0$)
\begin{equation}
\boldsymbol{s}^{(n)}_{f,1}=\mathcal{M}_{1:0}(\boldsymbol{s}^{(n)}_{0}),
\end{equation}
resulting in the ensemble of forecast results,
\begin{equation}
\boldsymbol{S}_{f, 1}=\left[\boldsymbol{s}^{(1)}_{f, 1},\boldsymbol{s}^{(2)}_{f, 1},\dots \boldsymbol{s}^{(n)}_{f, 1}, \dots \boldsymbol{s}^{(N-1)}_{f, 1},\boldsymbol{s}^{(N)}_{f, 1}\right].
\end{equation}
where $\boldsymbol{s}^{(n)}_{f,j+1}$ denotes the $n-\text{th}$ forecast member in the ensemble at time $t_{j+1}$ with a dimension of $D_s$. 
Afterwards the analysis step is initialized by evaluating the forecast mean 
\begin{equation}
\bar{\boldsymbol{s}}_{f,1} = \frac{1}{N} \Sigma_{n=1}^{N} \boldsymbol{s}^{(n)}_{f,1},
\end{equation}
and forecast error covariance matrix 
\begin{equation}
P_{f,1} = \mathcal{C}(\boldsymbol{S}_{f,1}),
\end{equation}
where $\mathcal{C}$ is a covariance operator defined as
\begin{equation}
    \mathcal{C}(\boldsymbol{S})=\frac{1}{N-1}\boldsymbol{S}' (\boldsymbol{S}')^{\text{T}},
\end{equation}
\begin{equation}
   \boldsymbol{S}'=[\boldsymbol{s}^{(1)}-\bar{s},\boldsymbol{s}^{(2)}-\bar{s},\dots \boldsymbol{s}^{(n)}-\bar{s},\dots \boldsymbol{s}^{(N-1)}-\bar{s},\boldsymbol{s}^{(N)}-\bar{s}].
\end{equation}
In the meantime, the corresponding noisy observation ${s}_{m,1}$ is available, as well as its error distribution information (i.e. covariance matrix $\boldsymbol{R}_1$). By assuming the noise is additive, the observation can be related to the state vector in the model space as
\begin{equation}
\boldsymbol{s}_{m,1}=\boldsymbol{H}\boldsymbol{s}_1+\xi_1,
\end{equation}
where $\xi_j$ is the observation error (with $j=1$ at the current step). $\boldsymbol{H}$ is the linear observation operator from the model space to the observation space. In this study, we restrict the discussion to the linear observation operator, while the proposed algorithm can be conveniently extended to the nonlinear case. Moreover, we assume that $\boldsymbol{s}_{m,j}$ can be full or sparse observation with a dimension of $D_\text{obs}$, and therefore $D_\text{obs} \leq D_{s}$.  Afterwards, the ensemble of measurements is produced as

\begin{equation}
\boldsymbol{S}_{m, 1}=\left[\boldsymbol{s}^{(1)}_{m, 1},\boldsymbol{s}^{(2)}_{m, 1},\dots \boldsymbol{s}^{(n)}_{m, 1}, \dots \boldsymbol{s}^{(N-1)}_{f, 1},\boldsymbol{s}^{(N)}_{m, 1}\right],
\end{equation}
\begin{equation}
\boldsymbol{s}^{(n)}_{m,1}=\boldsymbol{s}_{m,1}+\xi^{(n)}_1,
\end{equation}
where $\xi^{(n)}_1$ is randomly sampled based on $\boldsymbol{R}_1$. Finally, the analysis step is completed by combining the forecast and measurement as
\begin{equation}
\boldsymbol{S}_{a,1} = \boldsymbol{S}_{f,1} + \boldsymbol{K}_{1} \left( {\boldsymbol{S}}_{m,1} - \boldsymbol{H} {\boldsymbol{S}}_{f,1} \right),
\label{eq:enkf}
\end{equation}
where 
\begin{equation}
\boldsymbol{K}_{j} = \boldsymbol{P}_{f,j} \boldsymbol{H}^T \left(\boldsymbol{H}  \boldsymbol{P}_{f,j} \boldsymbol{H}^T + \boldsymbol{R}_j \right)^{-1}
\end{equation}
is the Kalman gain. Afterwards, both the forecast and analysis steps are repeated for all future time instants $t=t_2, t_3 ,\dots$, until certain criteria are satisfied (e.g. reaching the desired forecast time $t_{\text{max}}$).
\label{sec:coupledalgorithim}

\begin{figure}[h]
    \centering
\includegraphics[width=0.9\textwidth, trim=8cm 0cm 8cm 0cm, clip]{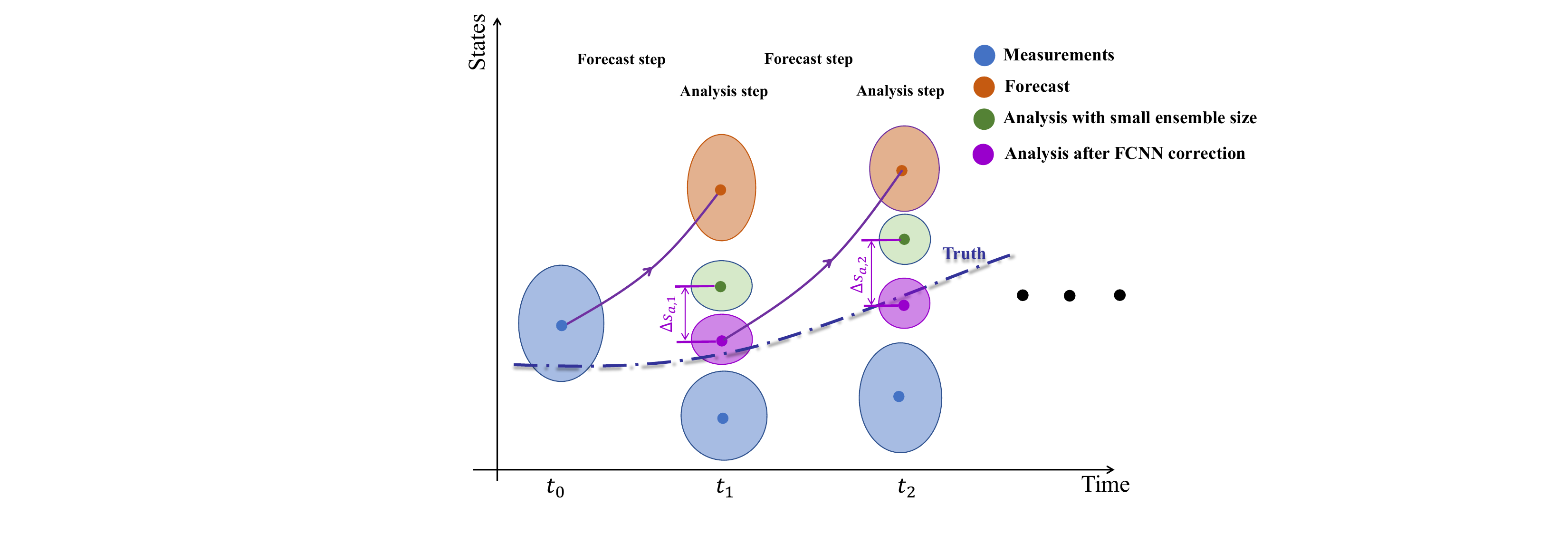}
    \caption{Schematic illustration of the EnKF-FCNN coupled algorithm. For the purpose of brevity, only the ensemble mean is shown, omitting individual members.}
    \label{fig:ENKFFCNN}
\end{figure}

\subsection{EnKF-FCNN coupled algorithm}
We consider a relatively small ensemble size ${\mathfrak{N}}$, with which the analysis results (specifically analysis mean $\bar{\boldsymbol{s}}^{\mathfrak{N}}_{a,j}$) while are better than the forecasts, but still show considerable deviations from the truth ${\boldsymbol{s}}_{t,j}$. Our goal is to utilize FCNN to build a function with the output being a correction term $\Delta{\boldsymbol{s}}_{a,j}$ as shown in Fig.~\ref{fig:ENKFFCNN}, and the final amended analysis mean given by the EnKF-FCNN algorithm $\bar{\boldsymbol{s}}^{\mathfrak{N}}_{a,j} \Leftarrow \bar{\boldsymbol{s}}^{\mathfrak{N}}_{a,j}+\Delta{\boldsymbol{s}}_{a,j}$ represents a more accurate approximation of ${\boldsymbol{s}}_{t,j}$. Given that the truth is usually unknown in practical applications, the analysis mean $\bar{\boldsymbol{s}}^{\mathcal{N}}_{a,j}$  given by a relatively large ensemble size $\mathcal{N}$ ($\gg\mathfrak{N}$) is assumed to be a precise surrogate. Therefore, the correction term, namely the output of FCNN, is defined as
\begin{equation}
\Delta{\boldsymbol{s}}_{a,j}=\bar{\boldsymbol{s}}^{\mathcal{N}}_{a,j}-\bar{\boldsymbol{s}}^{\mathfrak{N}}_{a,j}.
\label{eq:correct}
\end{equation} 
In addition, the input variables of the proposed FCNN-based function include $\boldsymbol{S}^{\mathfrak{N}}_{a,j}$, $\bar{\boldsymbol{s}}_{m,j}$, and  $\bar{\boldsymbol{s}}_{a,j-1}$. That is to say, our primary goal is to build a function via FCNN as
\begin{equation}
    \Delta{\boldsymbol{s}}_{a,j}=\mathcal{F}(\boldsymbol{S}^{\mathfrak{N}}_{a,j}, \bar{\boldsymbol{s}}_{m,j}, \bar{\boldsymbol{s}}_{a,j-1};\boldsymbol{\theta}),
\label{eq:func}
\end{equation}
where $\boldsymbol{\theta}$ represents the parameters. We remark that, although the EnKF formula Eq.~\eqref{eq:enkf} only includes the current forecast ensemble and the current observations (i.e., the previous state is not explicitly considered), its derivation depends on the flow-dependent assumption, namely the Markov chain assumption. Therefore, $\bar{\boldsymbol{s}}_{a,j-1}$ is included as one input to consider the influence of the immediately preceding state.

With the correction term $\Delta{\boldsymbol{s}}_{a,j}$ given by Eq.~\eqref{eq:func}, each ensemble member is updated as
 \begin{equation}
     {\boldsymbol{s}}^{(n)}_{a,j}\Leftarrow {\boldsymbol{s}}^{(n)}_{a,j}+\Delta{\boldsymbol{s}}_{a,j}, ~~n=1,2\dots\mathfrak{N},
 \end{equation}
before performing the next forecast step. Finally, a pseudocode for the complete EnKF-FCNN coupled algorithm is shown in Algorithm~\ref{al:ENKFFCNN}.

\begin{algorithm}
\caption{Algorithm for the EnKF-FCNN coupled method}
\begin{algorithmic}[1]
\State {\bf{Input}}: $\boldsymbol{S}_{0}^{\mathfrak{N}}$ (initial conditions); $\mathcal{F}$ (pretrained function)
\State {\bf{Begin}}:\\
initialize:\\
\hspace{1.2cm}  $t=t_0,~j=0$, $\boldsymbol{S}_{a,0}^{\mathfrak{N}}=\boldsymbol{S}_{m,0}^{\mathfrak{N}}$\\
time loop:
\State \hspace{1.2cm} {\bf{while}} $t \leq t_{\text{max}} $ {\bf{do}}\\
\hspace{1.6cm} $\boldsymbol{S}^{\mathfrak{N}}_{f,j+1}=\mathcal{M}_{j+1:j}(\boldsymbol{S}^{\mathfrak{N}}_{a,j})$\\
\hspace{1.6cm} Calculate $\boldsymbol{S}^{\mathfrak{N}}_{a,j+1}$ with~\eqref{eq:enkf}\\
\hspace{1.6cm} Calculate $\Delta{\boldsymbol{s}}_{a,j+1}$ with~\eqref{eq:func}\\
\hspace{1.6cm} Update $\boldsymbol{S}^{\mathfrak{N}}_{a,j+1}$ with ${\boldsymbol{s}}^{(n)}_{a,j+1}\Leftarrow {\boldsymbol{s}}^{(n)}_{a,j+1}+\Delta{\boldsymbol{s}}_{a,j+1}$, $n=1,2\dots\mathfrak{N}$\\
\hspace{1.6cm} {\bf{Output}} $\bar{\boldsymbol{s}}^{\mathfrak{N}}_{a,j}$ (corrected ensemble mean)\\
\hspace{1.6cm}  $j \Leftarrow j+1$; $t \Leftarrow t_{j}$\\
\hspace{1.2cm} {\bf{end}} \\
{\bf{end}} 
\end{algorithmic}
\label{al:ENKFFCNN}
\end{algorithm}

\section{Numerical Experiments and Results}
\label{sec:Experiment results}

To evaluate the feasibility of the proposed EnKF-FCNN algorithm and investigate the impact of various parameters (including ensemble size, observation site number, and observation frequency), we perform a set of numerical experiments by applying it in  Lorenz systems~\citep{lorenz1963deterministic,lorenz1996predictability} and the evolution of nonlinear ocean wave fields. Featured with the chaotic nature, the Lorenz systems are classic ``toy'' problems in the field of data assimilation; specifically, both Lorenz-63 and Lorenz-96 are tested in this study. In addition, the evolution of nonlinear ocean wave fields has also been shown to be highly sensitive to the model input and parameters, i.e. a small deviation, say, of initial conditions and physical parameters could induce significant forecast errors. Recent studies have shown that DA methods can be conveniently applied to realize the high-accuracy nonlinear wave field simulation~\citep{wang2021phase,wang2022phase,Liu2025}, which while depends on the relatively large ensemble size ($\sim\text{O}(10^2)$), ad-hoc operations (covariance inflation and localization), or both. Here we adopt the Pseudospectral-Fourier-Legendre (PFL) wave model to simulate the nonlinear wave field evolution and evaluate the EnKF-FCNN algorithm by extending the EnKF-PFL computational framework developed in~\citep{Liu2025}.

\subsection{Lorenz-63 system}
The Lorenz-63 system is expressed as
\begin{eqnarray}
\frac{d\chi}{dt} &=& \sigma \left( \upsilon - \chi \right), \label{eq:l631}\\
\frac{d\upsilon }{dt} &=& \chi (\rho - \zeta) - \upsilon,  \label{eq:l632}\\
\frac{d\zeta}{dt} &=& \chi\upsilon - \beta \zeta, \label{eq:l633}
\label{eq:Lorenz-63 differential}
\end{eqnarray}
where $\chi$, $\upsilon$, and $\zeta$ represent the system states.  $\sigma$, $\rho$, and $\beta$ are the model parameters, which are set to $10$, $28$, and $8/3$, respectively, in this study. To solve Lorenz-63 numerically, Eqs.~\eqref{eq:l631}-\eqref{eq:l633} are discretized in time with the Runge–Kutta fourth-order (RK4) scheme
and a time step of $0.01$ model time unit (MTU).

At the analysis step, the measurements are needed, which can approximate the truth yet with some noise. In this study, we first produce the (synthetic) truth $\boldsymbol{s}_{t,j}$ by running a set of reference simulations with presumed exact initial conditions. By following~\citep{lang2017data}, the exact initial condition is taken as the state after a spin-up period of $200$ MTU to avoid the transient effect. Then an error $\boldsymbol{\delta}$ randomly sampled from a given distribution is superposed on top of ${\boldsymbol{s}}_{t,j}$ to produce $\bar{\boldsymbol{s}}_{m,j}$, i.e.
\begin{equation}
\bar{\boldsymbol{s}}_{m,j}={\boldsymbol{s}}_{t,j}+\boldsymbol{\delta}.
\end{equation}
Here we assume that $\boldsymbol{\delta}$ follows a normal distribution with zero mean and covariance matrix $\boldsymbol{\Sigma}$. Particularly, we set $\boldsymbol{\Sigma}=A \boldsymbol{I}$, with $A$ being the magnitude (here $A = 2$) and $\boldsymbol{I}$ being an identity matrix. Finally, the observation ensemble members are generated as  
\begin{equation}
    \boldsymbol{s}^{(n)}_{m,j}= \bar{\boldsymbol{s}}_{m,j}+\boldsymbol{\delta}^{(n)}_e,
\end{equation}
where $\boldsymbol{\delta}^{(n)}_e$ is also a random error and assumed to follow the same distribution as $\boldsymbol{\delta}$. 

\begin{figure}
    \centering
    \begin{subfigure}[h]{0.32\textwidth}
        \centering   \includegraphics[width=\textwidth]{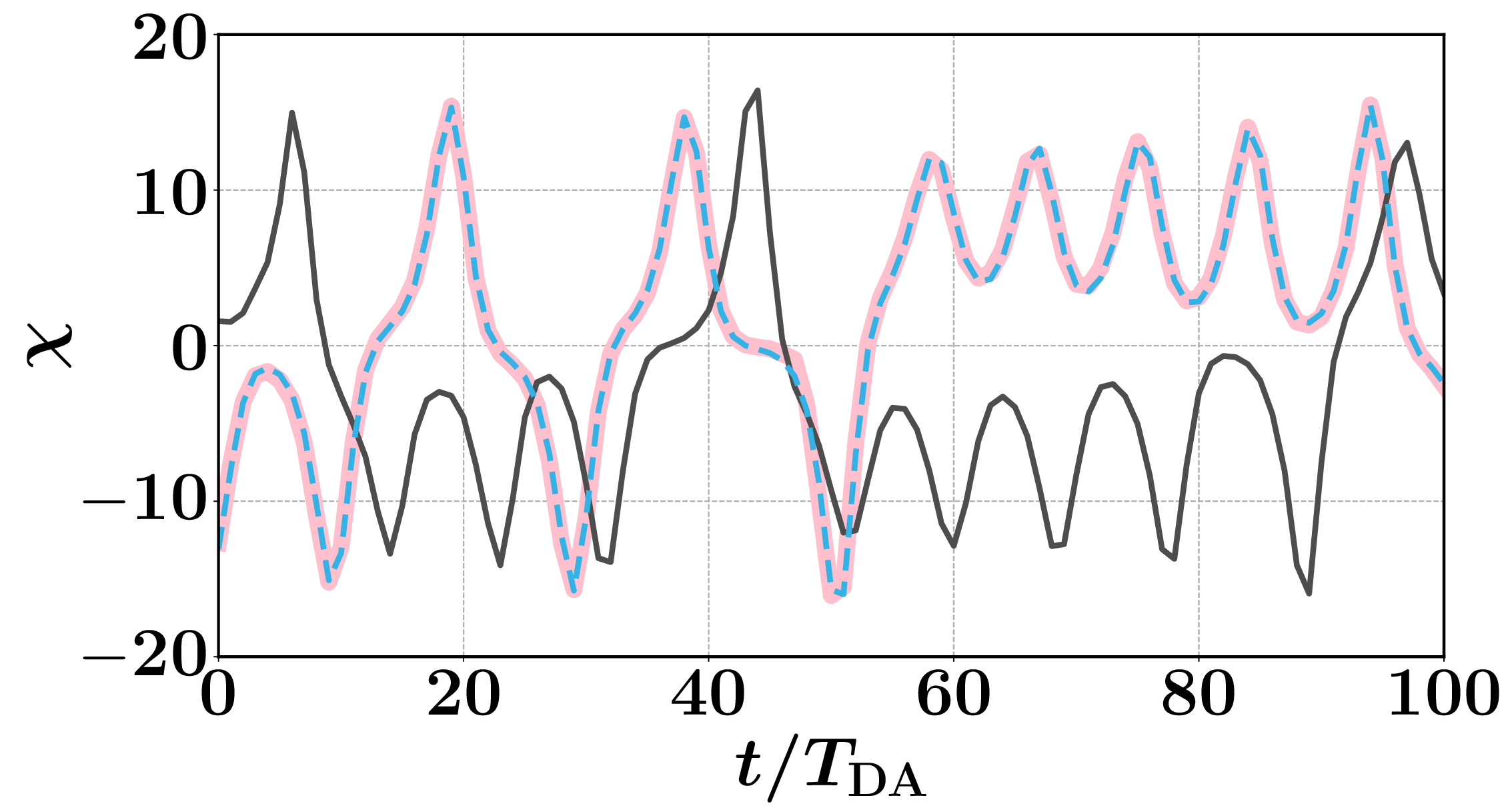}
        \caption{}  
        
    \end{subfigure}  
    \begin{subfigure}[h]{0.32\textwidth}
        \centering     \includegraphics[width=\textwidth]{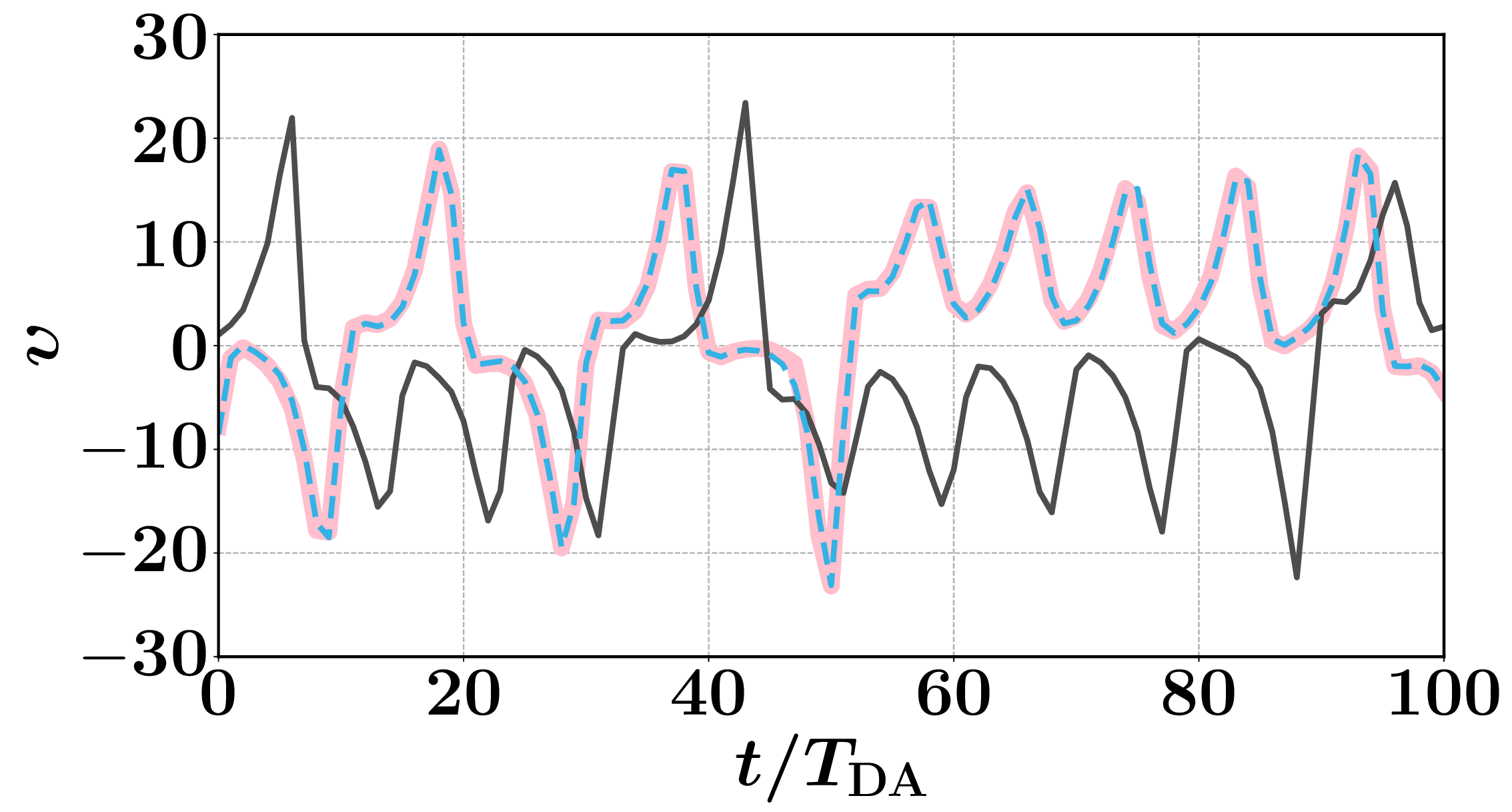}      \caption{}  
        
    \end{subfigure}
    \begin{subfigure}[h]{0.32\textwidth}
        \centering     \includegraphics[width=\textwidth]{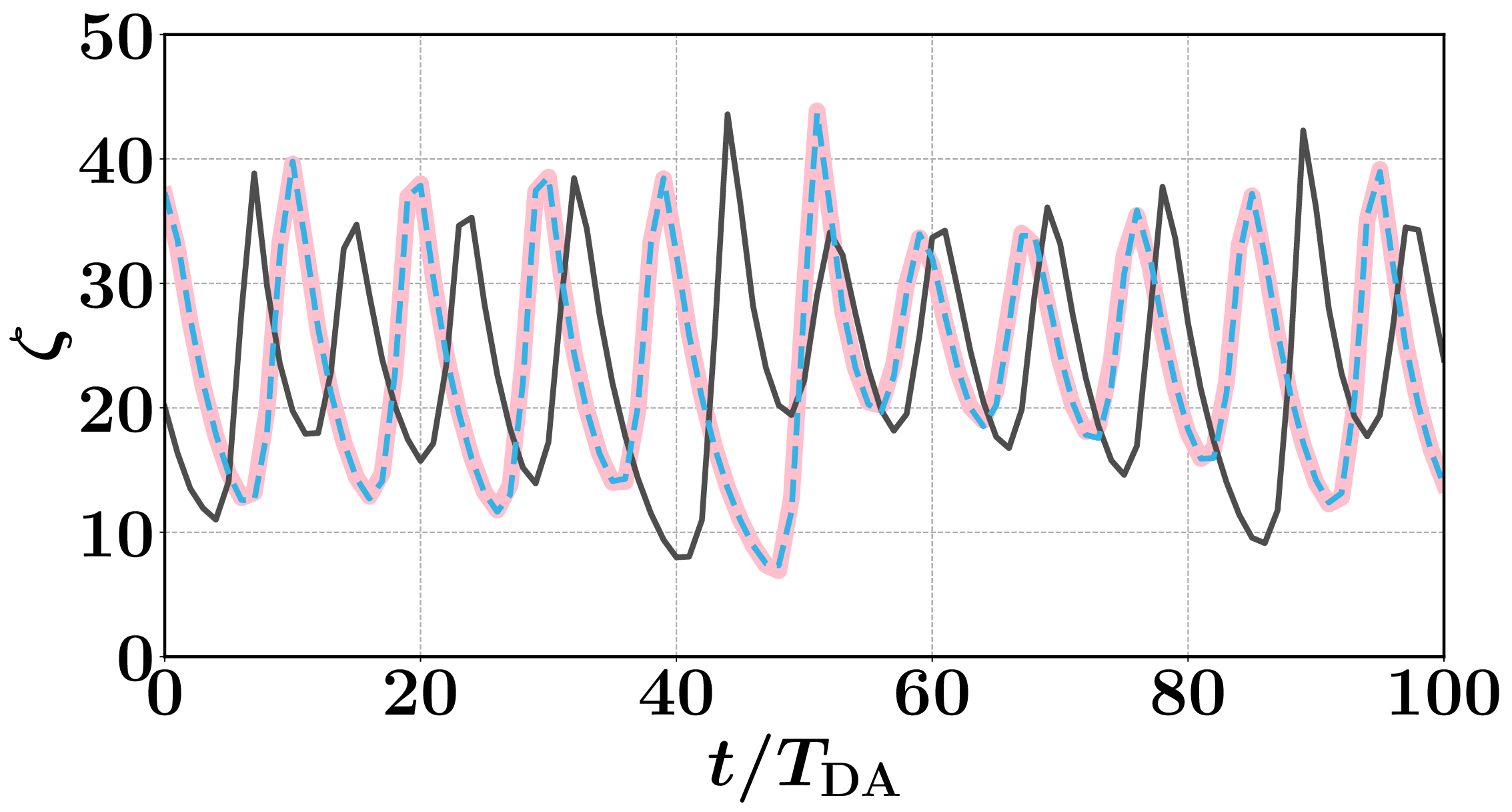}      \caption{}  
        
    \end{subfigure}
    \caption{Analysis results obtained with the traditional EnKF using $\mathcal{N}=100$ ({\color{mypink}\rule[0.5ex]{0.5cm}{2.0pt}}) and $\mathfrak{N}=3$ ({\color{darkgray}\rule[0.5ex]{0.5cm}{0.5pt}}), as well as the true solution ({\color{cyan}\dashL}), for Lorenz-63 benchmark case: (a)~$\chi$, (b)~$\upsilon$, and (c)~$\zeta$.}
     \label{L63state_xyz8dt}
 \end{figure}

To generate the datasets needed to build FCNN, we perform the traditional EnKF-based simulations (i.e. without FCNN) with $\mathcal{N}$ and $\mathfrak{N}$. Then the results are divided into training, validation, and test sets with a ratio of $70:15:15$. Here we set $\mathcal{N}=100$, with which the analysis results can accurately approximate the truth (see Fig.~\ref{L63state_xyz8dt}). In the training process, we use the rectified linear unit (ReLU) as the activation function and mean squared error (MSE) as the loss function, and the detailed configuration of other hyperparameters is shown in Tab.~\ref{tab:parameters}.

\begin{table}[h]
\centering
\caption{Hyperparameters of FCNN for different numerical simulations}
\label{tab:parameters}
\begin{tabular}{ccccc}
\toprule
 & \textbf{Lorenz-63} & \textbf{Lorenz-96} & \textbf{2D Wave} & \textbf{3D Wave}\\ \midrule
size of input layer  & \multicolumn{4}{c}{$D_{s}(\mathfrak{N}+1)+D_\text{obs}$} \\
size of hidden layer 1 & 60 & 200 &10000 &8000 \\
size of hidden layer 2 & 15 & 100 &2500 &4000\\
size of hidden layer 3 & 7 & 40 &500 &2000\\
size of output layer  & \multicolumn{4}{c}{$D_{s}$} \\
\bottomrule
\end{tabular}
\end{table}

\begin{figure}
    \centering
    \begin{subfigure}[h]{0.32\textwidth}
        \centering   \includegraphics[width=\textwidth]{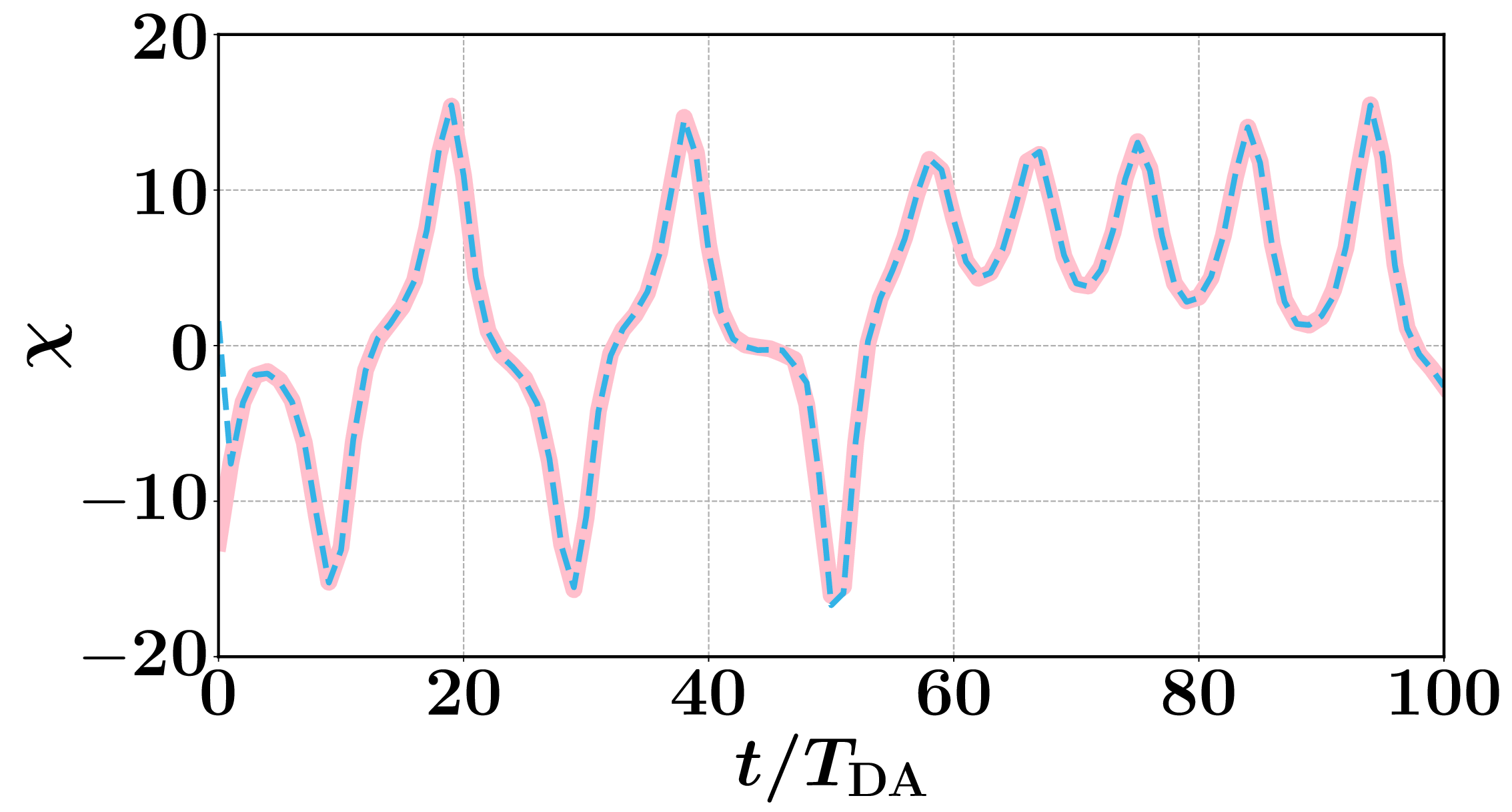}
        \caption{}  
      
    \end{subfigure}  
    \begin{subfigure}[h]{0.32\textwidth}
        \centering     \includegraphics[width=\textwidth]{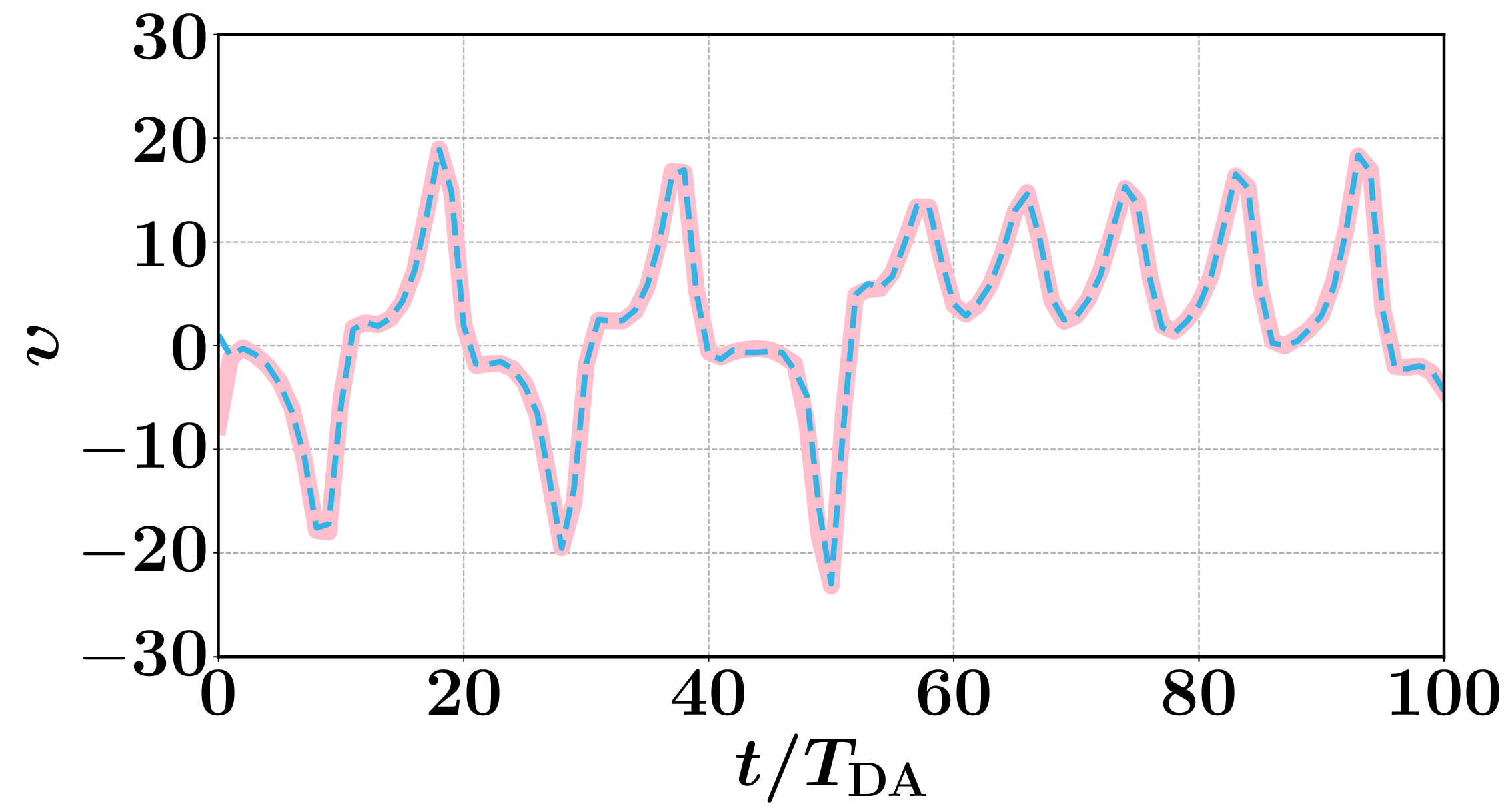}      \caption{}  
        
    \end{subfigure}
    \begin{subfigure}[h]{0.32\textwidth}
        \centering     \includegraphics[width=\textwidth]{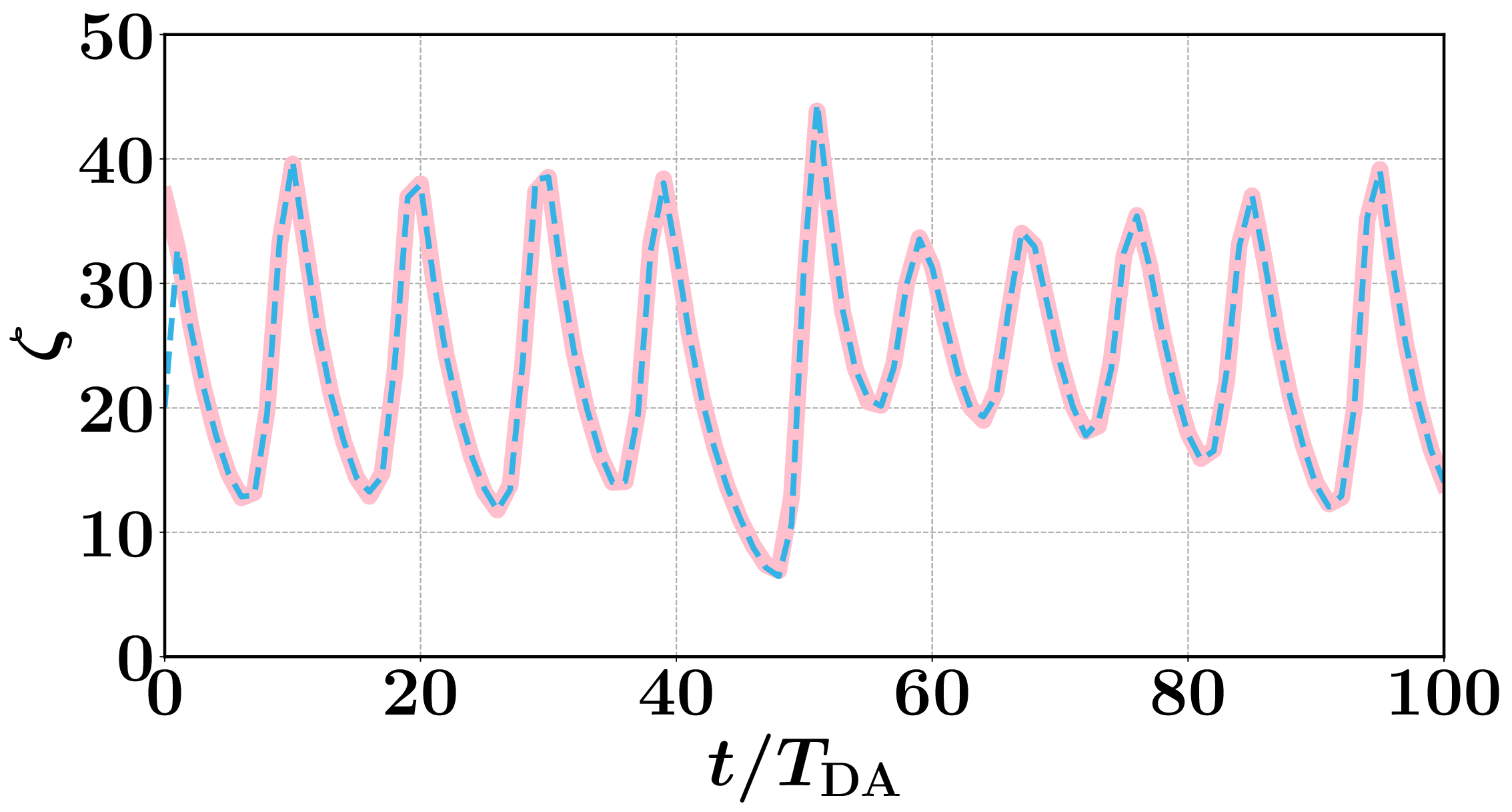}      \caption{}  
        
    \end{subfigure}
    \caption{Analysis results given by the EnKF-FCNN algorithm with $\mathfrak{N}=3$({\color{cyan}\dashL}) and traditional EnKF with $\mathcal{N}=100$ ({\color{pink}\rule[0.5ex]{0.5cm}{2.0pt}}) for Lorenz-63 benchmark case: (a)~$\chi$, (b)~$\upsilon$, and (c)~$\zeta$.}
     \label{L63time_xyz8dt}
 \end{figure}

We first perform a benchmark numerical experiment with $\mathfrak{N}=3$, DA interval $T_{\text{DA}}=0.08~\text{MTU}$, and all three variables observed. As shown in Fig.~\ref{L63state_xyz8dt}, the analysis results produced by the traditional EnKF for the benchmark case show significant discrepancies from the truth. Then at each analysis step, we add the FCNN-based correction term.  Fig.~\ref{L63time_xyz8dt} presents the results of $\bar{\boldsymbol{s}}^{\mathfrak{N}}_{a}$ produced by the EnKF-FCNN algorithm in comparison with $\bar{\boldsymbol{s}}^{\mathcal{N}}_{a}$ given by the traditional EnKF. It can be found that, with the FCNN-based correction term, $\bar{\boldsymbol{s}}^{\mathfrak{N}}_{a}$ can always closely follow $\bar{\boldsymbol{s}}^{\mathcal{N}}_{a}$, which indicates that the proposed EnKF-FCNN algorithm can effectively ameliorate the filter divergence issue suffered by the traditional EnKF in the scenario with a limited ensemble size.

\begin{figure}
    \centering
    \begin{subfigure}[h]{0.45\textwidth}
        \centering   \includegraphics[width=\textwidth]{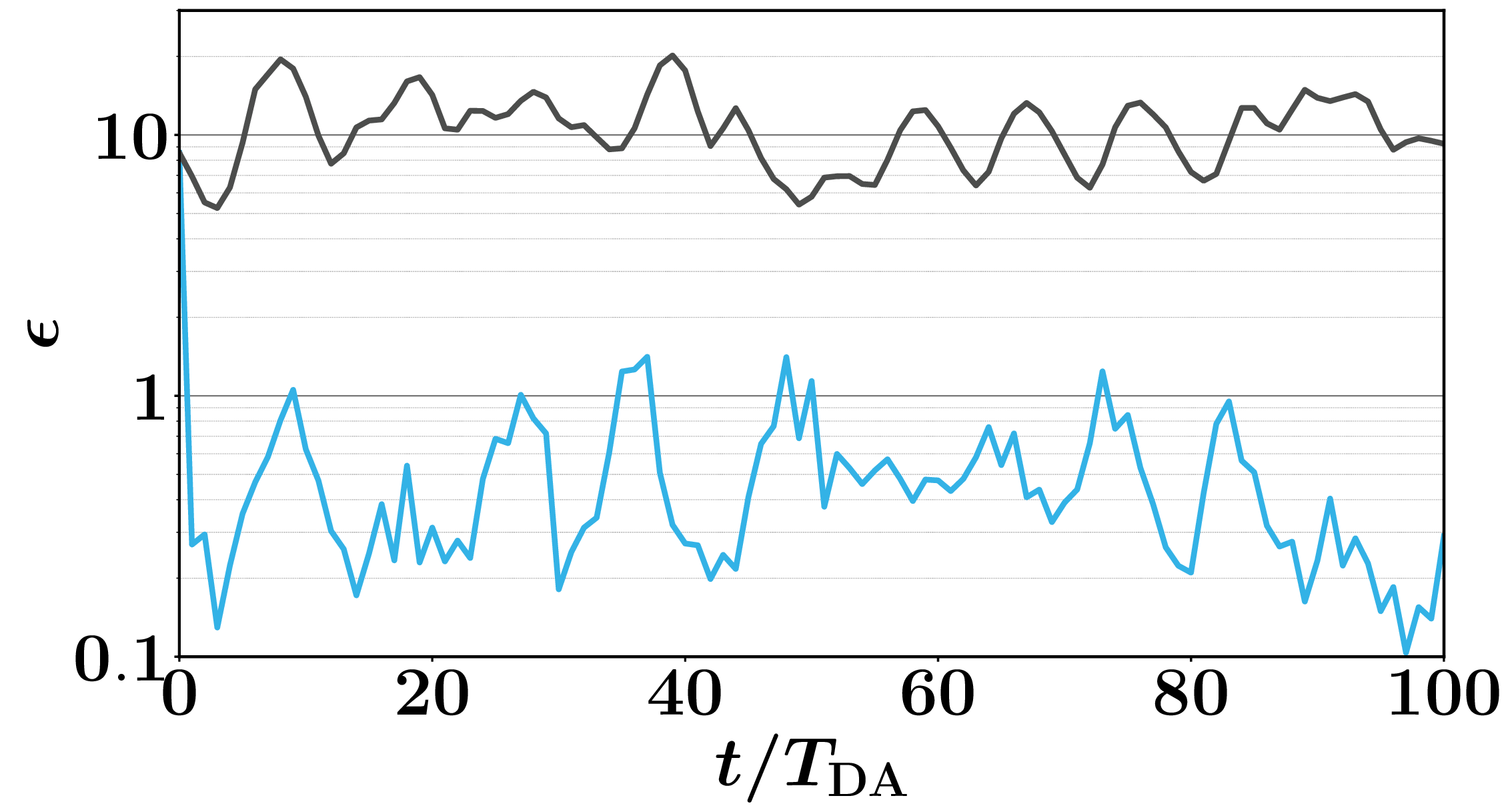}
        \caption{}  
        
    \end{subfigure}  
    \begin{subfigure}[h]{0.45\textwidth}
        \centering     \includegraphics[width=\textwidth]{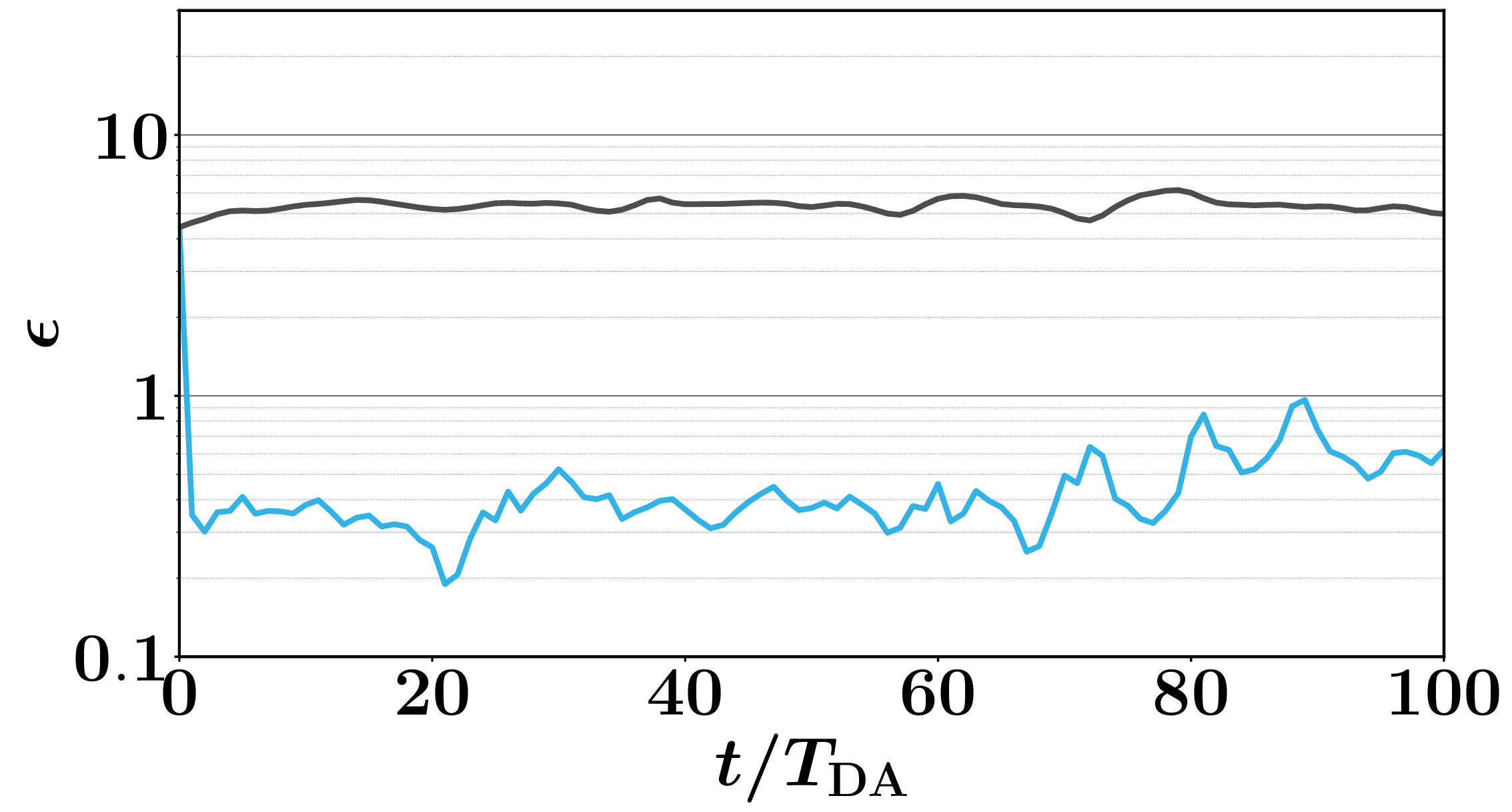}      \caption{}  
        
    \end{subfigure}
   \caption{Time histories of $\epsilon$ with traditional EnKF ({\color{darkgray}\rule[0.5ex]{0.5cm}{0.5pt}}) and EnKF-FCNN ({\color{cyan}\rule[0.5ex]{0.5cm}{0.5pt}}) for benchmark cases: (a) Lorenz-63 and (b) Lorenz-96}
    \label{fig:epsilonhis}
 \end{figure}

 In addition, an error metric, i.e. root mean square error (RMSE), is defined to quantitatively evaluate the overall performance of the proposed EnKF-FCNN algorithm
\begin{equation}    
\epsilon(t_j)=\sqrt{\frac{1}{K}\Sigma_{k=1}^{K}\left(\bar{\boldsymbol{s}}^{\mathfrak{N}}_{a,j,k}-\bar{\boldsymbol{s}}^{\mathcal{N}}_{a,j,k}\right)^{2}},
\label{eq:L2norm}
\end{equation}
where $K$ is the number of initial conditions in the test dataset, with $k$ being the index.  Here we evaluate $\epsilon(t_j)$ using both the traditional EnKF and EnKF-FCNN with the same limited ensemble size $\mathfrak{N}=3$, for which the results are shown in Fig.~\ref{fig:epsilonhis}(a). It can be found that, after applying the FCNN-based correction, $\epsilon$ is reduced by more than one order of magnitude. Furthermore, we evaluate the computational time of the FCNN function, as compared to that of performing a single simulation for $T_\text{DA}$. As shown in Tab.~\ref{tab:CPU-time}, for Lorenz-63, the computational time consumed by implementing the FCNN-based correction is approximately one order of magnitude lower than running a single simulation for $T_\text{DA}$, which means that the additional computational cost induced by the former is practically negligible.

\begin{table}[h]
\centering
\caption{Computational time for different calculations on a single Intel Core Ultra 9 CPU}
\label{tab:CPU-time}
\begin{tabular}{ccc}
\toprule \textbf{Cases}
 & \textbf{A single simulation for $\boldsymbol{T}_\text{DA}$ (s)} & \textbf{FCNN function (s)}\\ \midrule
Lorenz-63 & 6.15e-4 & 7.68e-5
 \\
Lorenz-96 & 2.62e-3
 & 2.91e-4
 \\
2D wave & 4.69e-1
 & 8.62e-4
\\
3D wave & 2.72
 & 1.90e-3\\
\bottomrule
\end{tabular}
\end{table}

\begin{figure}
    \centering
    \begin{subfigure}[h]{0.45\textwidth}
        \centering   \includegraphics[width=\textwidth]{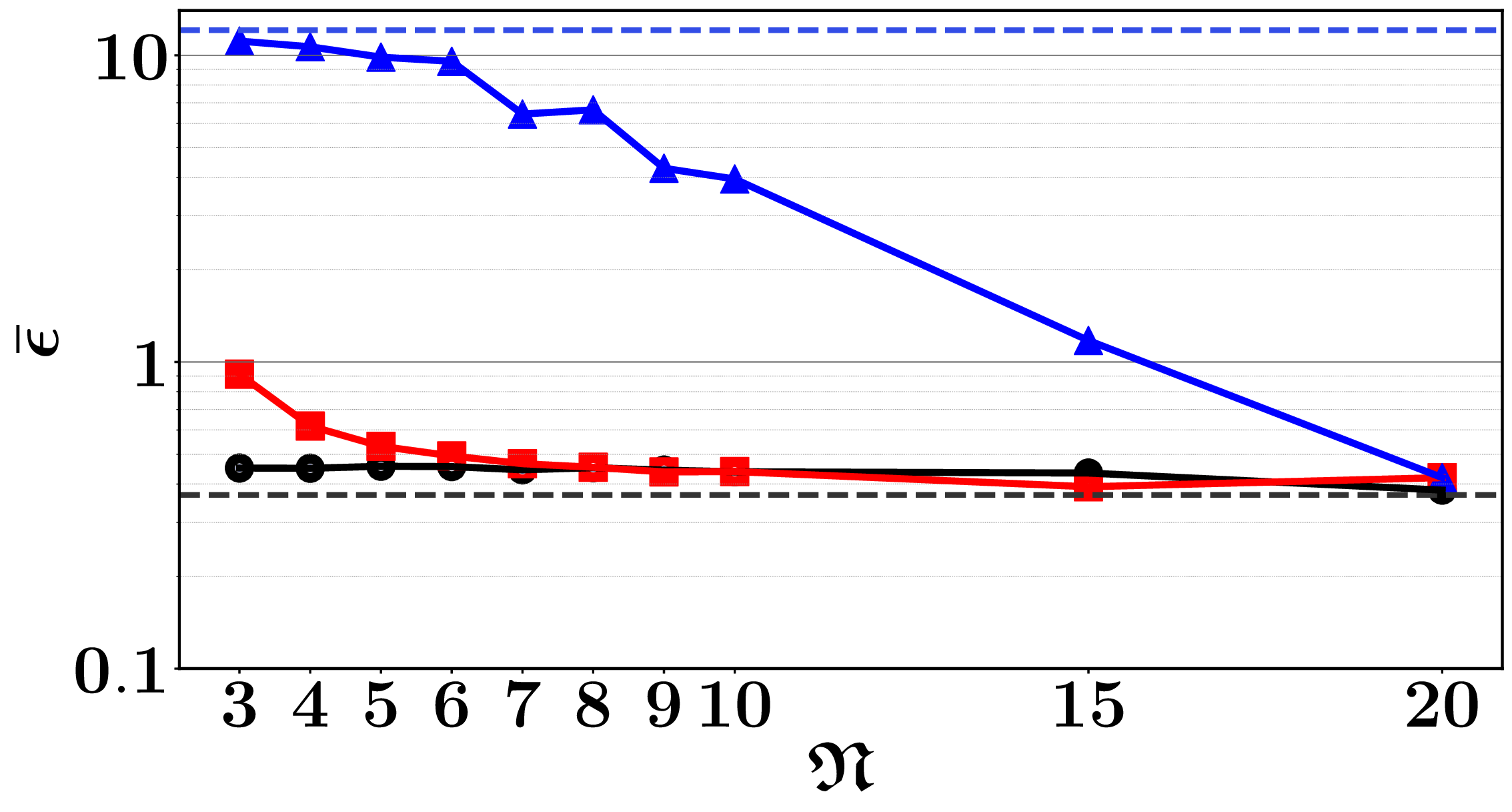}
        \caption{}  
        
    \end{subfigure}  
    \begin{subfigure}[h]{0.45\textwidth}
        \centering     \includegraphics[width=\textwidth]{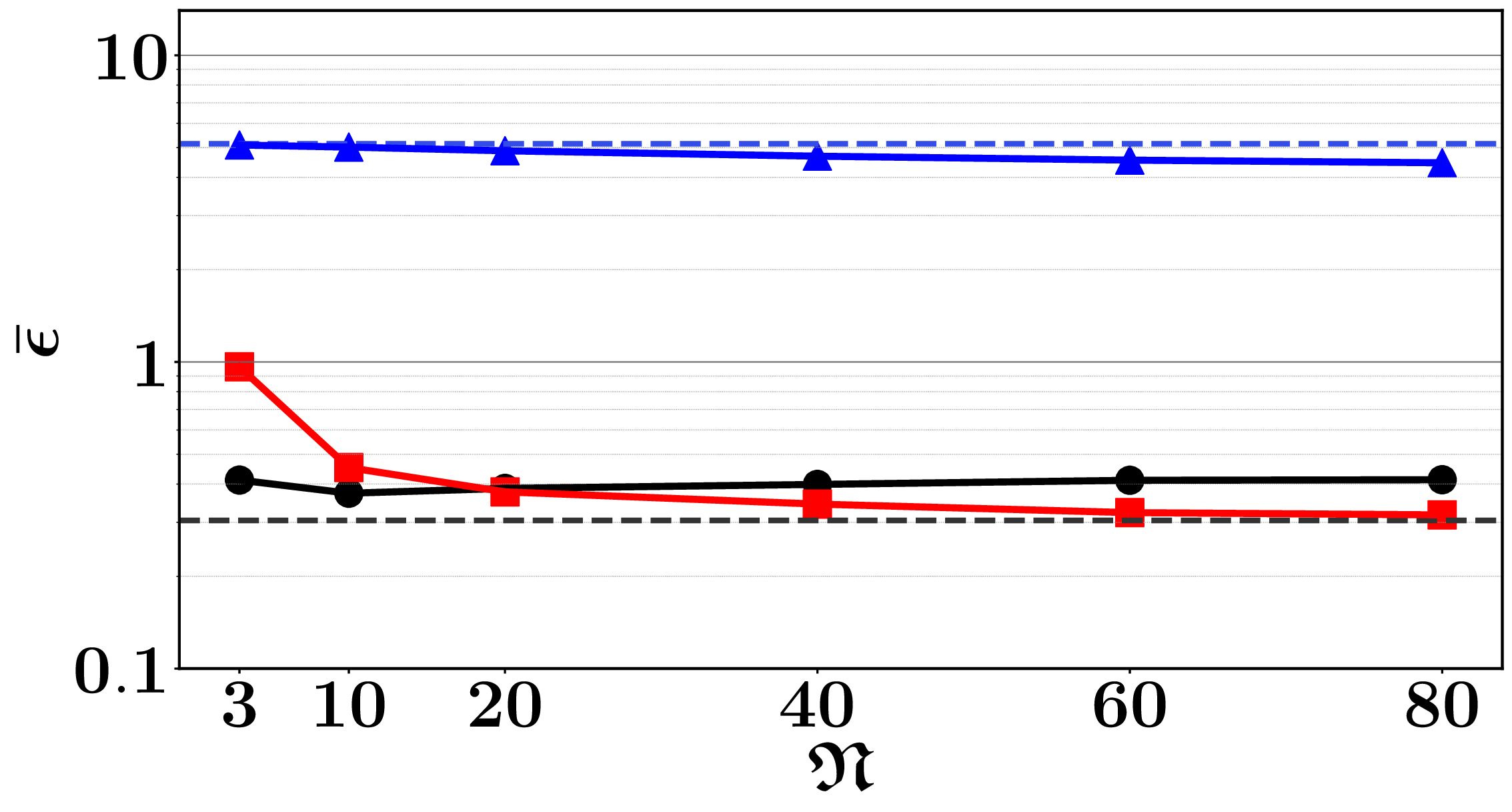}      \caption{}  
        
    \end{subfigure}
\caption{$\bar{\mathfrak{\epsilon}}$ given by the EnKF-FCNN algorithm (\rule[0.5ex]{0.5cm}{0.5pt} with circles), and traditional EnKF with ({\color{red}\rule[0.5ex]{0.5cm}{0.5pt}} with squares) and without ({\color{blue}\rule[0.5ex]{0.5cm}{0.5pt}} with triangles) the optimal inflation factor, for different values of $\mathfrak{N}$: (a) Lorenz-63 and (b) Lorenz-96.~{\color{blue}\dashL}~and~{\color{darkgray}\dashL}~represent $\bar{\mathfrak{\epsilon}}$ given by the traditional EnKF with $\mathcal{N}=100$ and without DA (i.e. free run), respectively.}
     \label{fig:epsilon}
 \end{figure}

\begin{figure}
    \centering
    \begin{subfigure}[h]{0.45\textwidth}
        \centering   \includegraphics[width=\textwidth]{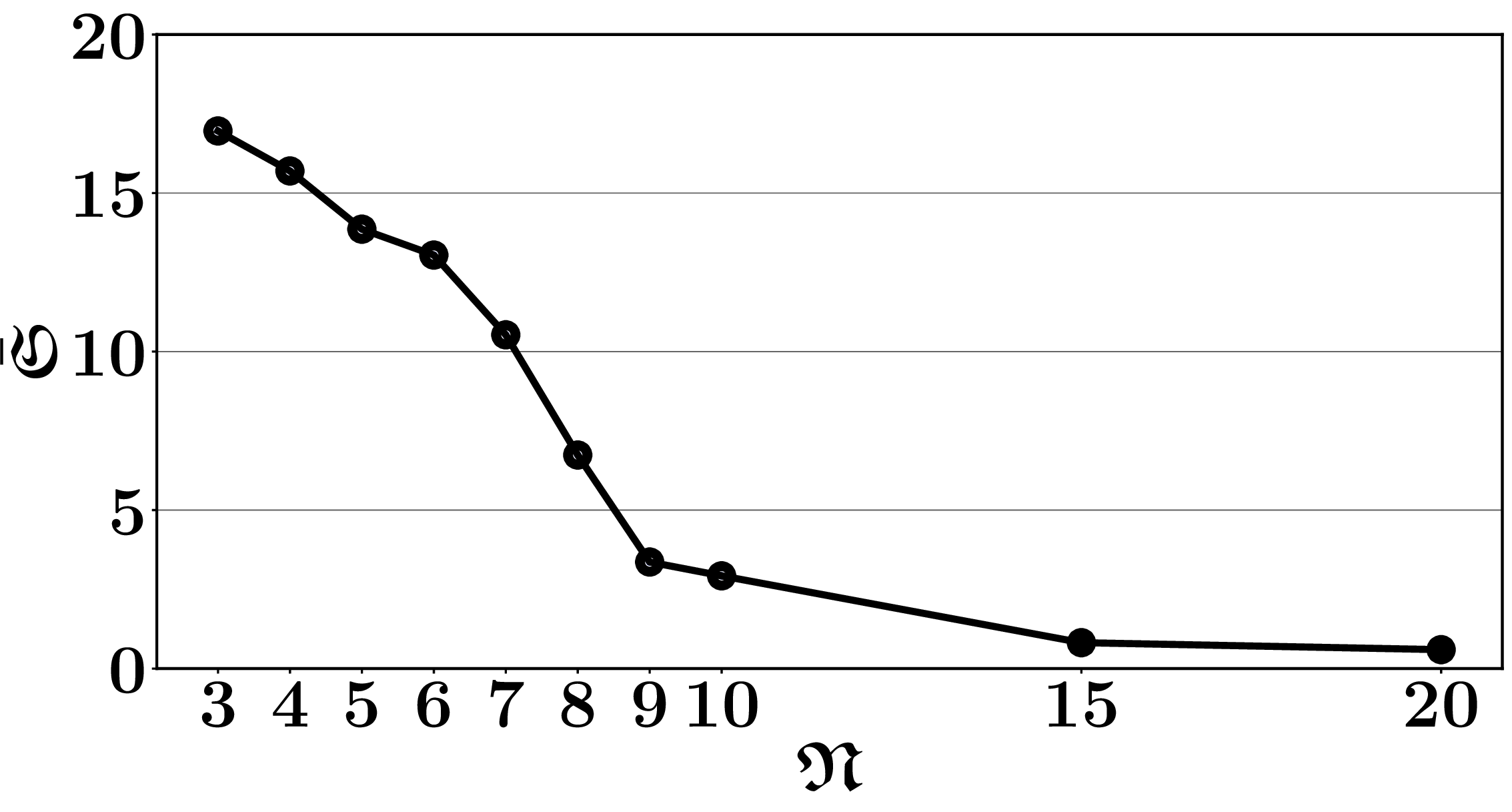}
        \caption{}  
        
    \end{subfigure}  
    \begin{subfigure}[h]{0.45\textwidth}
        \centering     \includegraphics[width=\textwidth]{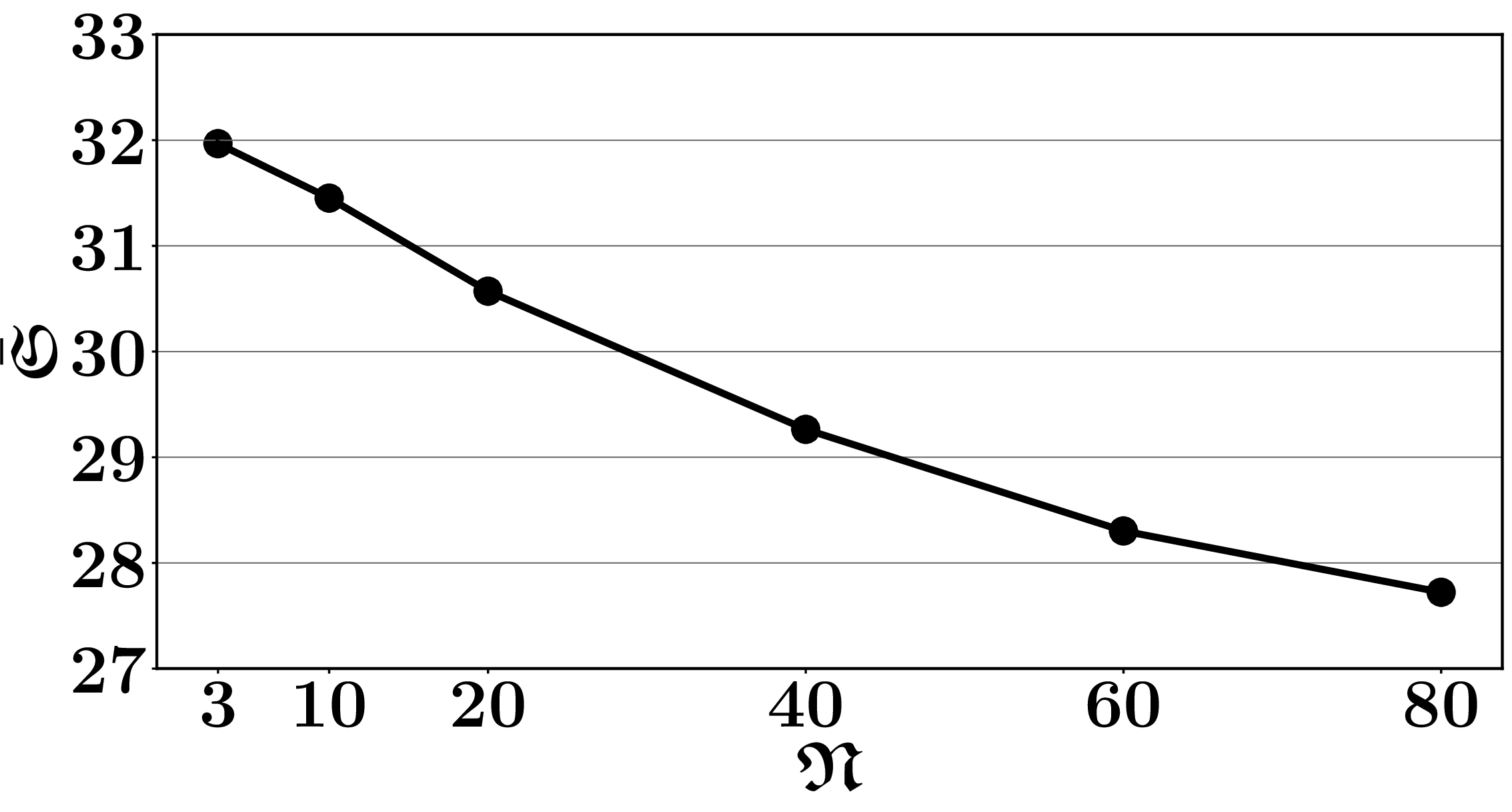}      \caption{}  
        
    \end{subfigure}
\caption{$\bar{\mathfrak{S}}$ with the EnKF-FCNN algorithm obtained with different values of $\mathfrak{N}$: (a) Lorenz-63 and (b) Lorenz-96}
     \label{fig:mag}
 \end{figure}

Afterwards, we further demonstrate the feasibility of the proposed EnKF-FCNN algorithm with different values of $\mathfrak{N}$, while all other parameters remain the same as the benchmark case. Specifically, for each $\mathfrak{N}$, we evaluate the time-averaged RMSE 
\begin{equation}
    \bar{\epsilon}=\frac{1}{\mathcal{K}_t}\Sigma_{j=1}^{\mathcal{K}_t}\epsilon(t_j)
\end{equation}
where $\mathcal{K}_t$ is the number of time instances. In addition, it should be noted that covariance inflation can also be a potential pathway to address the issues induced by a limited ensemble size~\citep{kang2012estimation}, and therefore it is necessary to compare the results of the proposed EnKF-FCNN to those produced by applying one optimal inflation factor. In this regard, for each $\mathfrak{N}$, we determine the optimal inflation factor ($\Lambda_\text{opt}$) through exhaustive search with the goal of minimizing $\bar{\epsilon}$, which are summarized in Tab.~\ref{tab:optinfl63}. Fig.~\ref{fig:epsilon} compares the results of $\bar{\epsilon}$ given by the traditional EnKF with and without the optimal inflation factor, for $\mathfrak{N}=3,~4,~5,~6,~7,~8,~9,~10,~15,~\text{and} ~20$, as well as those produced by the EnKF-FCNN algorithm. It is noted that for the tested range of $\mathfrak{N}$, $\bar{\epsilon}$ maintains a constant level ($\sim\text{O}(0.5)$) with slight oscillations and is always lower than that given by the traditional EnKF without the optimal inflation factor. Moreover, it can be found that, when $\mathfrak{N}$ is extremely small (e.g. $\mathfrak{N}=3$), the EnKF-FCNN algorithm can still significantly outperform the traditional EnKF even if the latter is enhanced by the optimal inflation factor, while their performances are comparable for other relatively larger ensemble sizes. In addition, to investigate the impact of $\mathfrak{N}$ on the magnitude of the correction term, we quantify the time-averaged magnitude of $\Delta{\boldsymbol{s}}_{a,j}$ (denoted as $\bar{\mathfrak{S}}$), for which the results are shown in Fig.~\ref{fig:mag}. It can be found that, when ${\mathfrak{N}}\leq10$, $\bar{\mathfrak{S}}$ decreases quasi-linearly with ${\mathfrak{N}}$; and $\bar{\mathfrak{S}}$ would converge gradually when ${\mathfrak{N}}$ is further increased.

Furthermore, we evaluate the impacts of the observation site number on the overall performance of the proposed EnKF-FCNN algorithm though a set of numerical experiments by assuming different combinations of available observations, including $\{\chi,~\upsilon,~\zeta\}$, $\{\chi,~\upsilon\}$, $\{\chi,~\zeta\}$, and $\{\chi\}$. All other settings still remain the same as the benchmark case. As shown in Tab.~\ref{tab:dt63}, $\bar{\epsilon}$ increases monotonically as we reduce the number of available observations. Moreover, it is notable that observations of $\{\chi,~\upsilon\}$ gives a better analysis accuracy compared to $\{\chi,~\zeta\}$, since $~\zeta$ can only provide the information of the height of the trajectory without constraining the other two coordinates $\{\chi,~\upsilon\}$~\citep{shen2023review}. In addition, $\bar{\mathfrak{S}}$ is also evaluated, which remains at the same level ($\sim\text{O}(16.5)$) for different observation combinations. One possible reason is that both two terms on the right-hand-side of Eq.~\ref{eq:correct} would change simultaneously as we switch the observations, and the magnitude of their difference (i.e. $\Delta{\boldsymbol{s}}_{a,j}$) is not quite sensitive to observation settings. Moreover, the influence of DA frequency is examined by testing two values of $T_{\text{DA}}=0.08~\text{and}~0.25~\text{MTU}$. As shown in Tab.~\ref{tab:TDA63}, a smaller value of $T_\text{DA}$ would lead to lower $\bar{\epsilon}$, while $\bar{\mathfrak{S}}$ is quite similar for both cases, for which the underlying reason is similar to that for observation combinations.

\begin{figure}
    \centering
    \begin{subfigure}[h]{0.45\textwidth}
        \centering   \includegraphics[width=\textwidth]{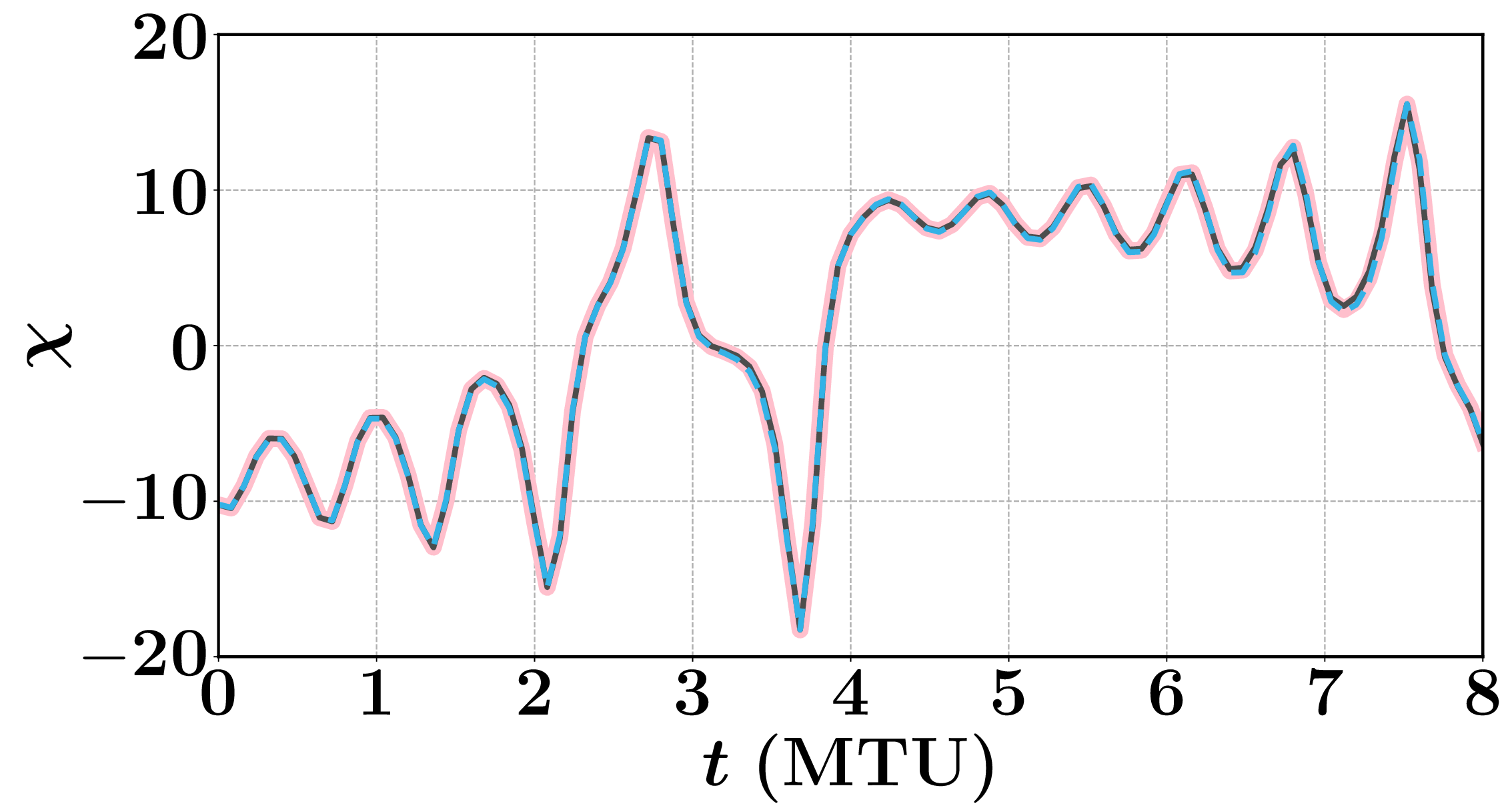}
        \caption{}  
        
    \end{subfigure}  
    \begin{subfigure}[h]{0.45\textwidth}
        \centering     \includegraphics[width=\textwidth]{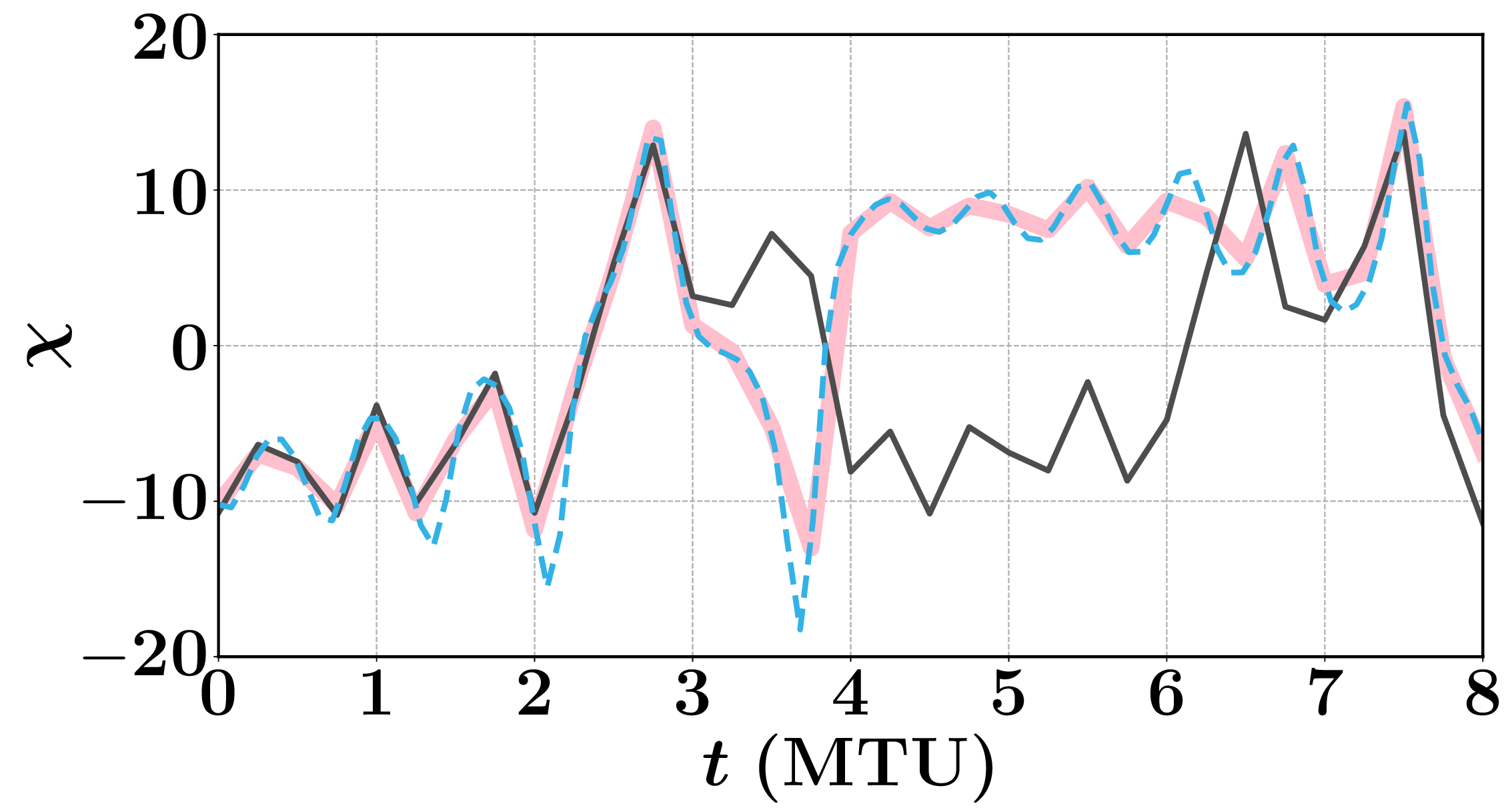}      \caption{}  
        
    \end{subfigure}
    \caption{Comparison of the results obtained with NN-DA ({\rule[0.5ex]{0.5cm}{1.0pt}}) and EnKF-FCNN ({\color{cyan}\dashL}) for (a) $T_\text{DA}=0.08~\text{MTU}$ and (b) $T_\text{DA}=0.25~\text{MTU}$, as well as the truth({\color{pink}\rule[0.5ex]{0.5cm}{2.0pt}})}
   \label{fig:compare}
\end{figure}

Finally, to demonstrate the necessity of keeping EnKF in the proposed algorithm, we compare the EnKF-FCNN algorithm with the method developed in~\citep{cintra2016tracking}, which replaces the traditional DA by a neural network (NN) completely and is denoted as NN-DA in the following text. Specifically, the NN-DA method builds one NN function to predict the analysis with the current forecast and observation as input. The training data are taken as the mean of analysis, forecast, and observation ensembles, which are produced from a simulation with EnKF using a sufficiently large ensemble size (e.g. $100$). The comparison is performed for both two DA intervals,  $T_\text{DA}=0.08~\text{and}~0.25~\text{MTU}$, with all other settings identical to the benchmark case. As shown in Fig.~\ref{fig:compare}, for  $T_\text{DA}=0.08~\text{MTU}$, NN-DA and EnKF-FCNN exhibit the comparable performance, both of which could closely follow the truth. However, for after increasing DA interval to $T_\text{DA}=0.25~\text{MTU}$, it is found that NN-DA shows significant discrepancies from the truth, while EnKF-FCNN can still work out well. One possible reason is that the input variables of NN cannot provide effective information about the covariance, which varies more significantly with time when a greater $T_\text{DA}$ is adopted. In contrast, as one input in EnKF-FCNN, the forecast ensemble can provide some (although maybe rough) insights into the temporal change of covariance, with which the overall framework can still remain effective for greater $T_\text{DA}$.  

\begin{table}[h]
\centering
\caption{Summary of optimal inflation factors for Lorenz-63}
\label{tab:optinfl63}
\begin{tabular}{lllllllllll}
\toprule
\textbf{$\mathfrak{N}$} & $3$ & $4$ & $5$ & $6$ & $7$ & $8$ & $9$ & $10$ & $15$ & $20$  \\ 
$\Lambda_{opt}$   & 1.57 & 1.19 & 1.13 & 1.08 & 1.06& 1.05  & 1.04 & 1.04 & 1.02 & 1.0 \\
\bottomrule
\end{tabular}
\end{table}

\begin{table}[h]
\centering
\caption{Summary of $\bar{\epsilon}$ and $\bar{\mathfrak{S}}$ with different observations for Lorenz-63}
\label{tab:dt63}
\begin{tabular}{lllll}
\toprule
\textbf{Observations} & $\{\chi,~\upsilon,~\zeta\}$ &  $\{\chi,~\upsilon\}$ & $\{\chi,~\zeta\}$ & $\{\chi\}$ \\ 
$\bar{\epsilon}$   & 0.44 & 0.59 & 0.68 & 1.18 \\
$\bar{\mathfrak{{S}}}$   & 17.01 & 16.49 & 16.56 & 16.71 \\
\bottomrule
\end{tabular}
\end{table}

\begin{table}[h]
\centering
\caption{Summary of $\bar{\epsilon}$ and $\bar{\mathfrak{S}}$ with different $T_\text{DA}$ for Lorenz-63}
\label{tab:TDA63}
\begin{tabular}{ccc}
\toprule
$T_\text{DA}$ & $0.08~\text{MTU}$ & $0.25~\text{MTU}$ \\ 
$\bar{\epsilon}$   & 0.44 & 0.80 \\
$\bar{\mathfrak{{S}}}$   & 17.01 & 16.25  \\
\bottomrule
\end{tabular}
\end{table}

\begin{table}[h]
\centering
\caption{Summary of optimal inflation factors and localization radii for Lorenz-96}
\label{tab:parameters96}
\begin{tabular}{ccccccccc}
\toprule
\textbf{$\mathfrak{N}$} & $3$ & $10$ & $20$ & $40$ & $60$ & $80$ \\ 
$\Lambda_{opt}$   & 1.18 & 1.06 & 1.04 & 1.03 & 1.02& 1.01\\
$\mathcal{R}_{opt}$   & 1.2 & 5 & 7 & 12 & 16& 18\\
\bottomrule
\end{tabular}
\end{table}

\begin{table}[h]
\centering
\caption{Summary of $\bar{\epsilon}$ and $\bar{\mathfrak{S}}$ with different observations for Lorenz-96}
\label{tab:obs96}
\begin{tabular}{cccc}
\toprule
\textbf{Observations} & $\{\chi^{(i)}~|~i=1:1:40\}$ & $\{\chi^{(i)}~|~i=1:2:39\}$ & $\{\chi^{(i)}~|~i=1:4:37\}$ \\ 
$\bar{\epsilon}$   & 0.24 & 0.37 & 0.90 \\
$\bar{\mathfrak{{S}}}$   & 30.59 & 30.92 & 31.79 \\
\bottomrule
\end{tabular}
\end{table}

\begin{table}[h]
\centering
\caption{Summary of $\bar{\epsilon}$ and $\bar{\mathfrak{S}}$ with different $T_\text{DA}$ for Lorenz-96}
\label{tab:TDA96}
\begin{tabular}{llll}
\toprule
$T_\text{DA}$ & $0.05~\text{MTU}$ & $0.10~\text{MTU}$ & $0.20~\text{MTU}$  \\ 
$\bar{\epsilon}$   & 0.37 & 1.23 & 1.96 \\
$\bar{\mathfrak{{S}}}$   & 30.92 & 30.38 & 29.75 \\
\bottomrule
\end{tabular}
\end{table}

\begin{figure}
    \centering
    \begin{subfigure}[h]{0.32\textwidth}
        \centering   \includegraphics[width=\textwidth]{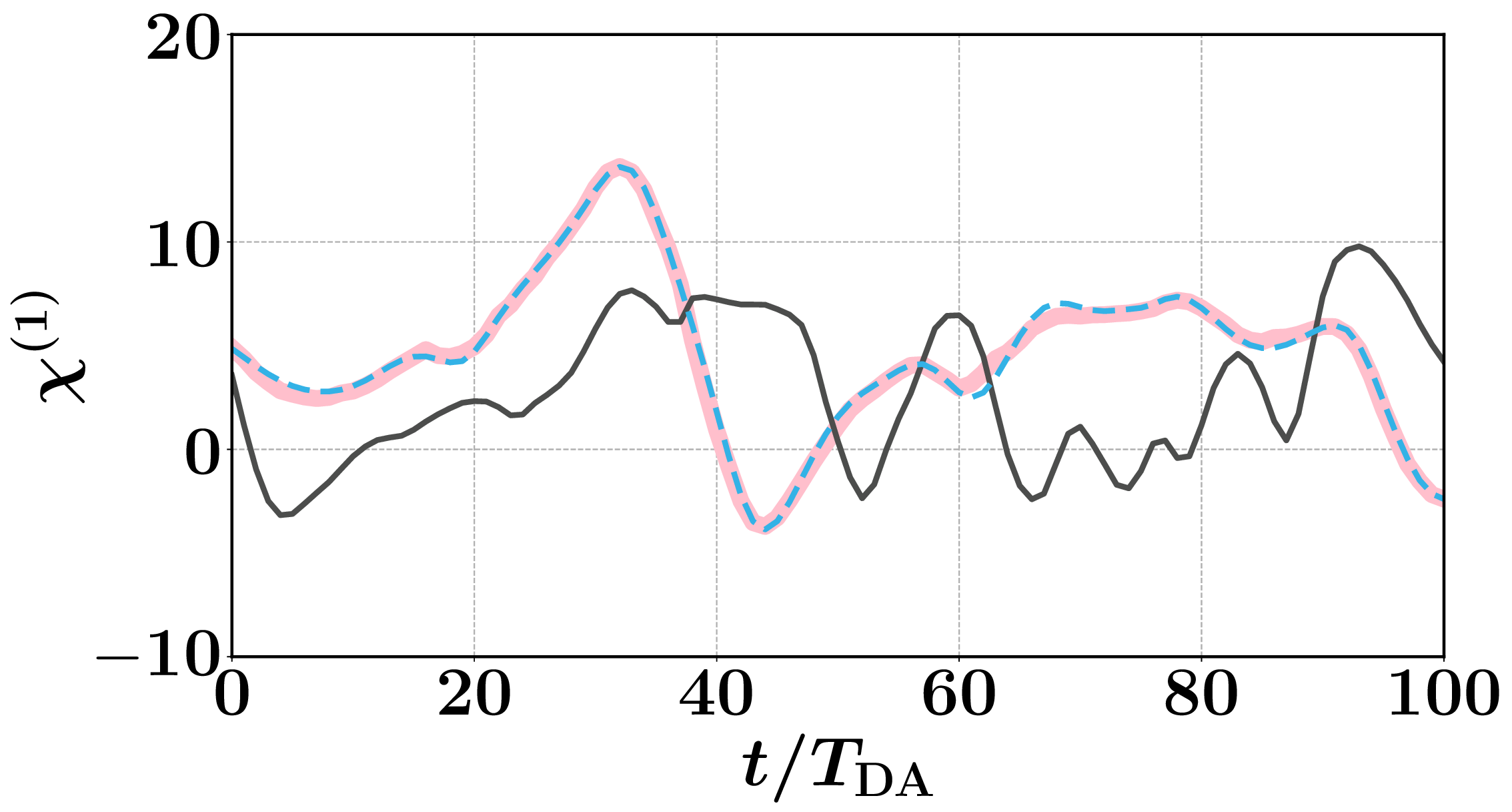}
        \caption{}  
        
    \end{subfigure}  
    \begin{subfigure}[h]{0.32\textwidth}
        \centering     \includegraphics[width=\textwidth]{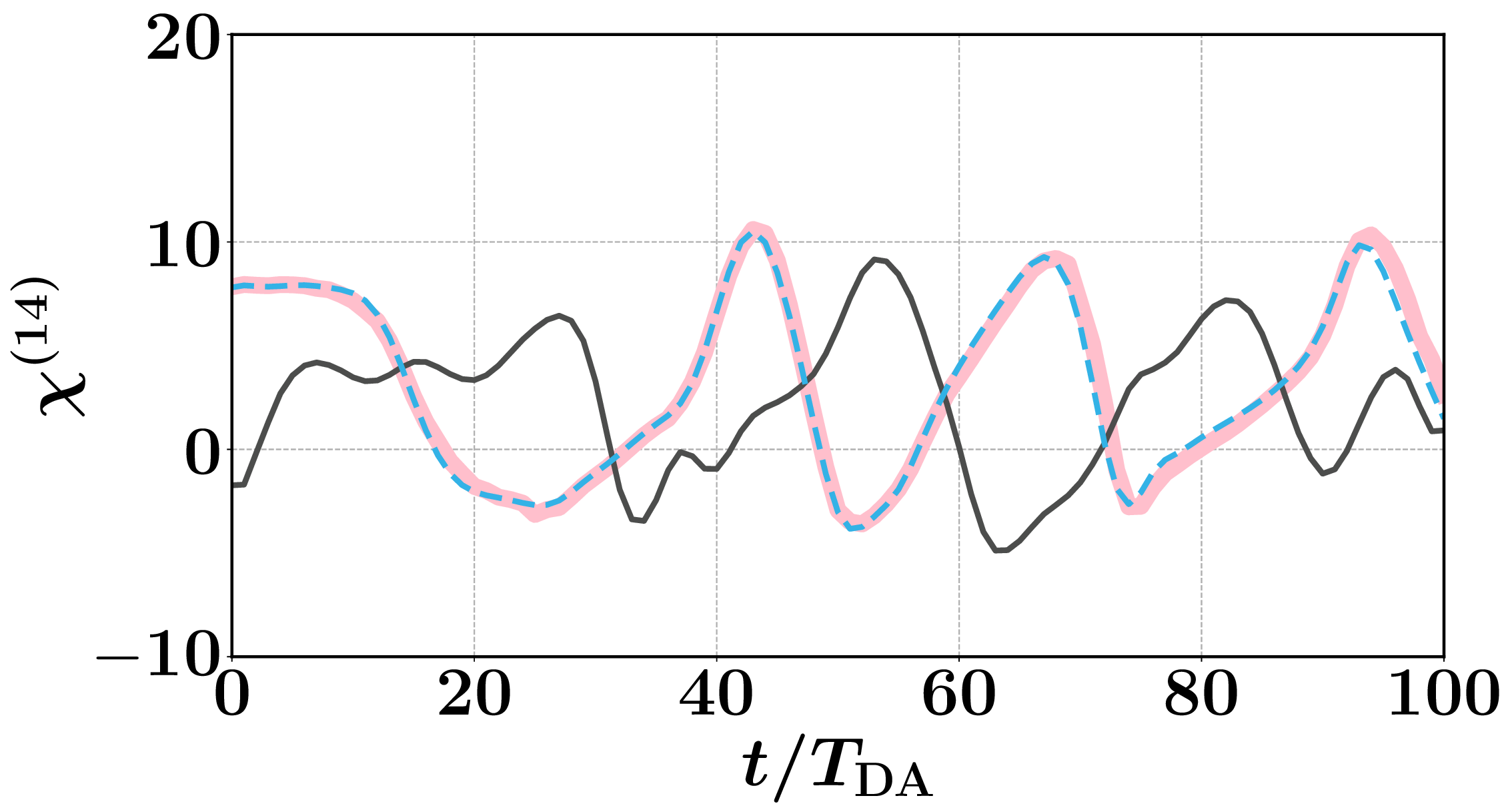}      \caption{}  
        
    \end{subfigure}
    \begin{subfigure}[h]{0.32\textwidth}
        \centering     \includegraphics[width=\textwidth]{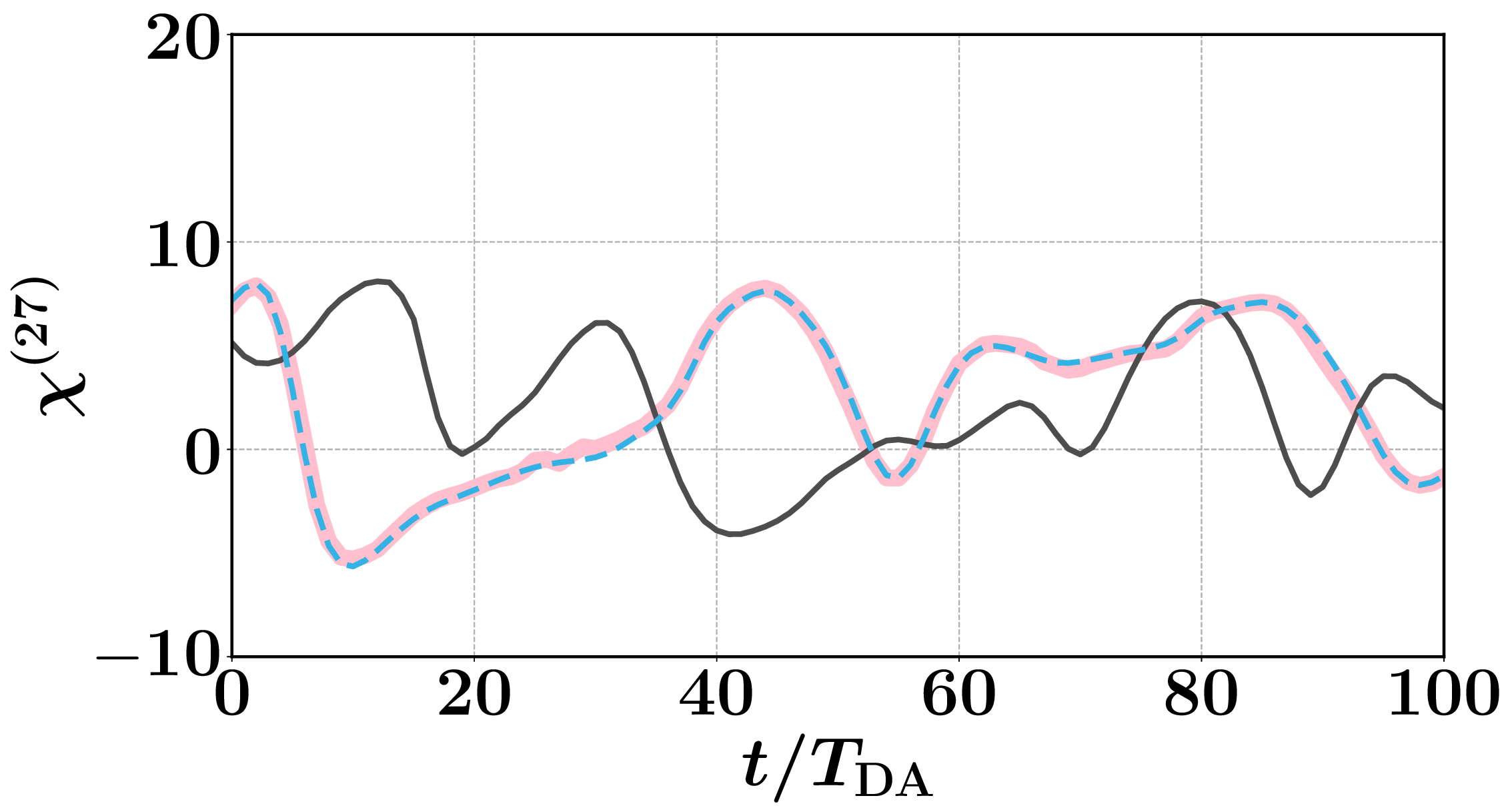}      \caption{}  
        
    \end{subfigure}
    \caption{Analysis results obtained with the traditional EnKF using $\mathcal{N}=100$ ({\color{mypink}\rule[0.5ex]{0.5cm}{2.0pt}}) and $\mathfrak{N}=10$ ({\color{darkgray}\rule[0.5ex]{0.5cm}{0.5pt}}), as well as the true solution ({\color{cyan}\dashL}), for Lorenz-96 benchmark case: (a)~$\chi^{(1)}$, (b)~$\chi^{(14)}$, and (c)~$\chi^{(27)}$.}
     \label{fig:L96state}
\end{figure}

\begin{figure}
    \centering
    \begin{subfigure}[h]{0.32\textwidth}
        \centering   \includegraphics[width=\textwidth]{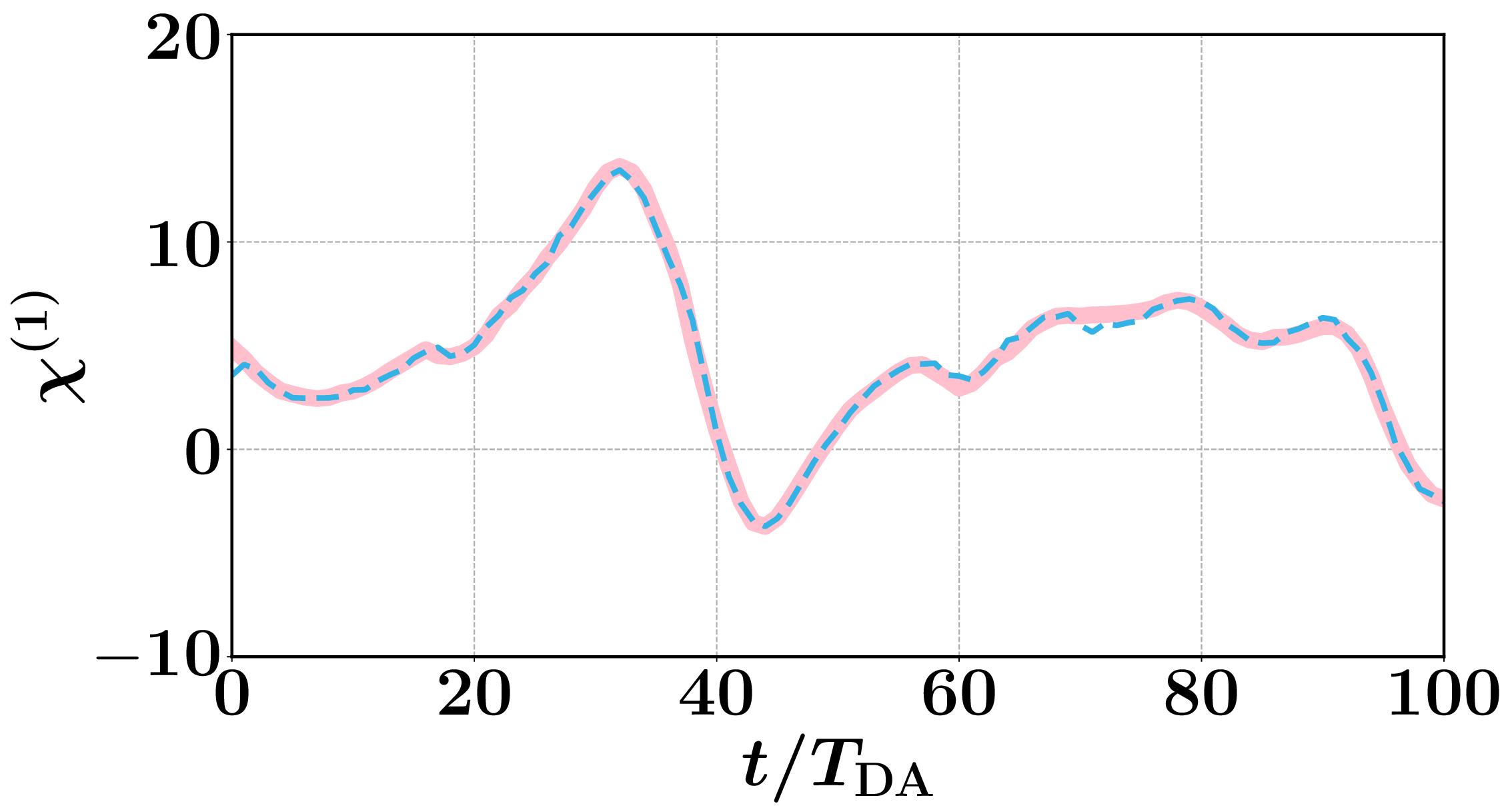}
        \caption{}  
        
    \end{subfigure}  
    \begin{subfigure}[h]{0.32\textwidth}
        \centering     \includegraphics[width=\textwidth]{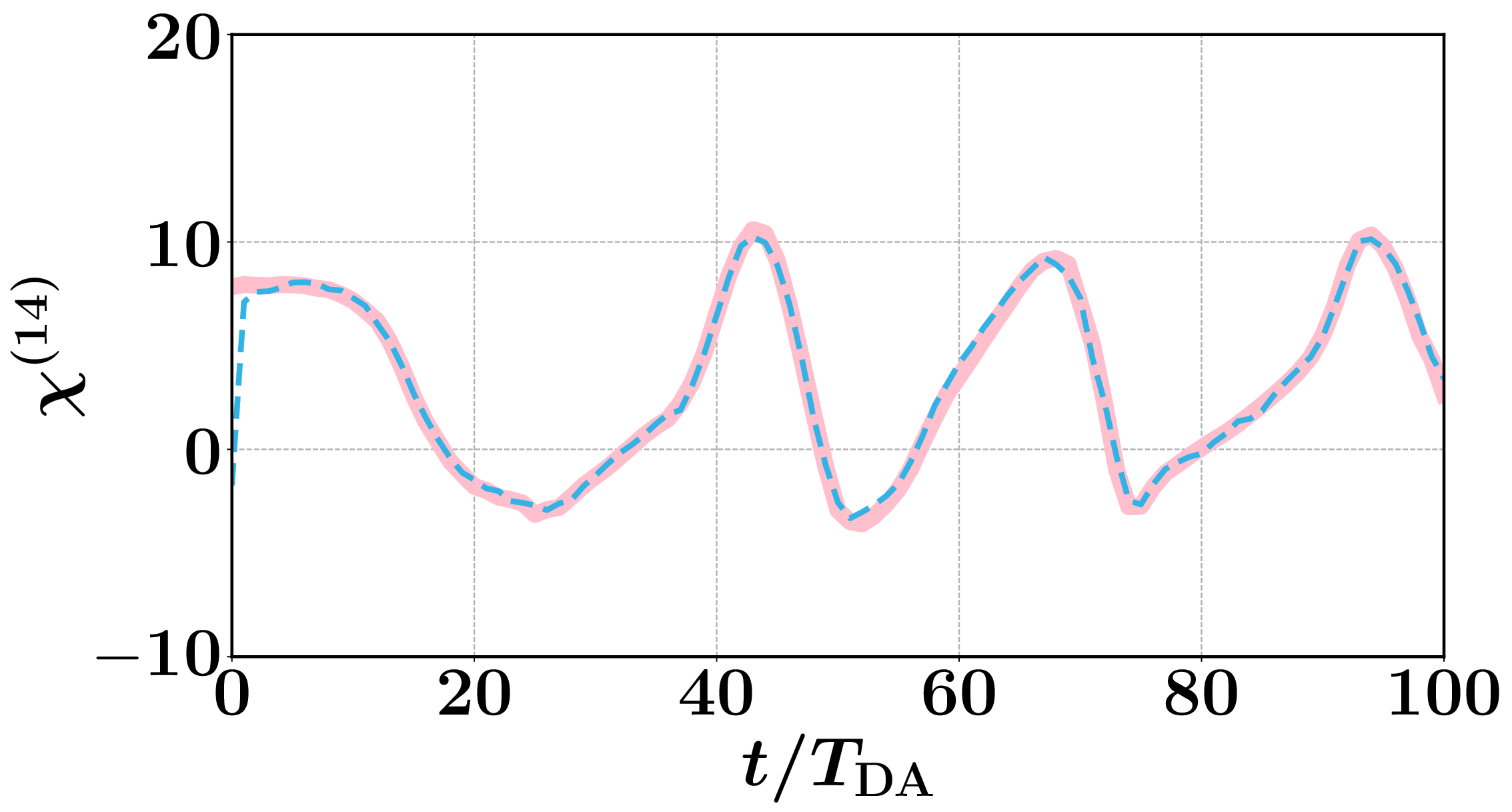}      \caption{}  
        
    \end{subfigure}
    \begin{subfigure}[h]{0.32\textwidth}
        \centering     \includegraphics[width=\textwidth]{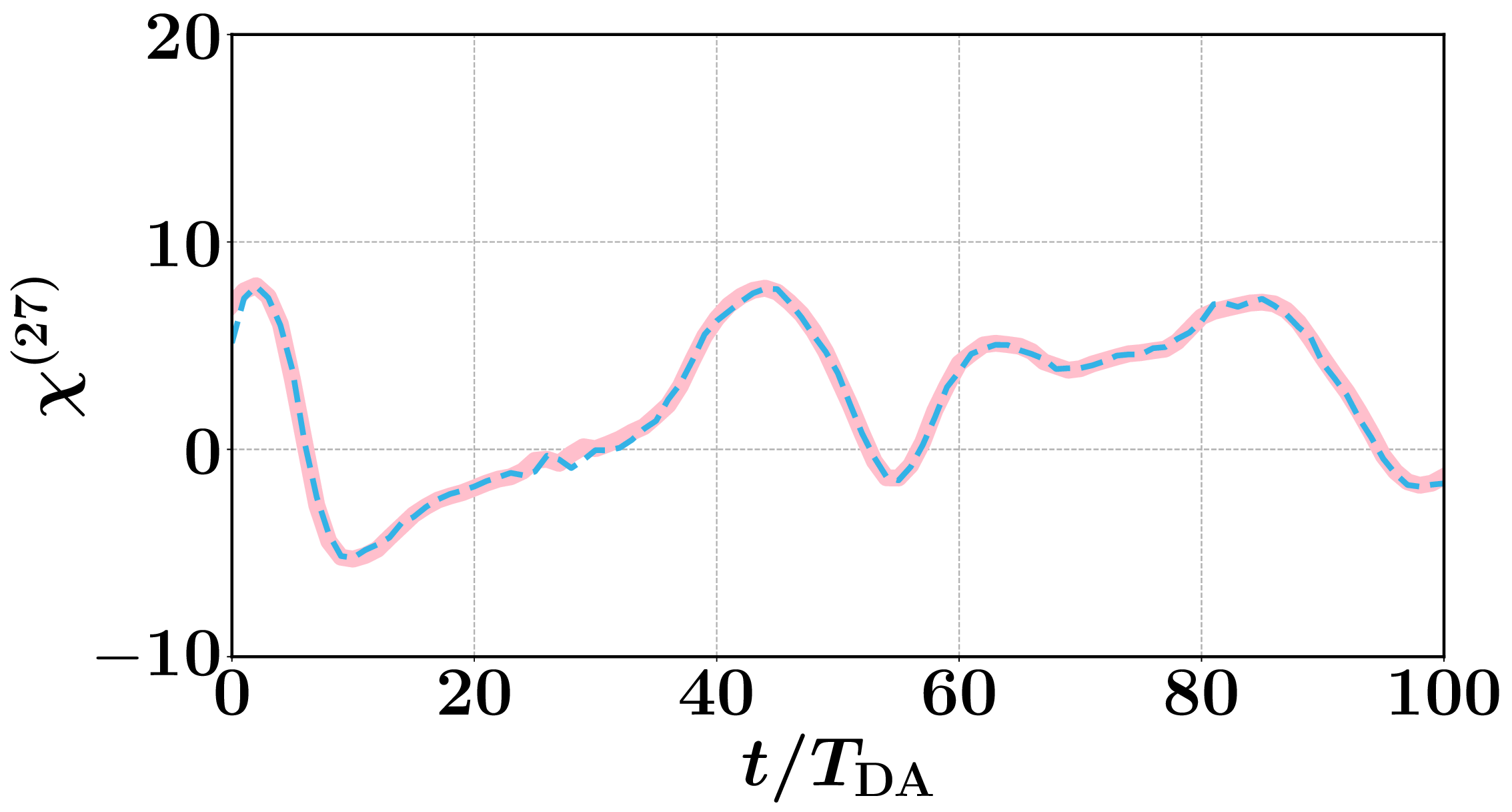}      \caption{}  
        
    \end{subfigure}
    \caption{Analysis results given by the EnKF-FCNN algorithm with $\mathfrak{N}=10$({\color{cyan}\dashL}) and traditional EnKF with $\mathcal{N}=100$ ({\color{pink}\rule[0.5ex]{0.5cm}{2.0pt}}) for Lorenz-96 benchmark case: (a)~$\chi^{(1)}$, (b)~$\chi^{(14)}$, and (c)~$\chi^{(27)}$.}
\label{fig:L96time}
\end{figure}

\subsection{Lorenz-96 system}
To further evaluate the scalability and robustness of the proposed method, we extend our experiments to the Lorenz-96 system, which is formulated as
\begin{equation}
\label{eq:Lorenz-96 differential}
\frac{d\chi^{(i)}}{dt} = (\chi^{(i+1)} - \chi^{(i-2)}) \chi ^{(i-1)} - \chi^{(i)} + F, i=1,2\dots L,
\end{equation}
where $\chi^{(i)}$ represents the $i-{\text{th}}$ state variable with $L$ variables in total. In this study, we consider a case of $L=40$, and the periodic boundary condition is applied.~$F$ is a constant external forcing term, which is set to be $F=8$. We still use the RK4 scheme and a time step of $0.01~\text{MTU}$ to solve Eq.~\eqref{eq:Lorenz-96 differential}. For the benchmark case, we consider $A=1$, $\mathfrak{N}=10$, $T_{\text{DA}}=0.05~\text{MTU}$, and $20$ available observations, i.e. $\{\chi^{(i)}|~i=1:2:39\}$. Here we still use $\mathcal{N}=100$ to define the correction term in Eq.~\eqref{eq:correct}, while incorporating localization to eliminate the spurious correlation and ensure the numerical stability. Specifically, the Gaspari–Cohn (GC) function~\citep{gaspari1999construction} is used to define the smooth correlation function
    \begin{equation}
    [\bm{\mu}]_{ij}=\begin{cases}1-\frac{5}{3}r^2+\frac{5}{8}r^3+\frac{1}{2}r^4-\frac{1}{4}r^5
    ~&\text{for}~0\leq r < 1\\4-5r+\frac{5}{3}r^2+\frac{5}{8}r^3-\frac{1}{2}r^4+\frac{1}{12}r^5-\frac{2}{3r}~&\text{for}~1\leq r < 2   \\0~&\text{for}~r\geq 2
    \end{cases}
,
\end{equation}
where $r={|i-j|}/{\mathcal{R}}$ and $\mathcal{R}$ is the localization radius. Then the regularized forecast covariance matrix, $\bm{\mu}\odot\boldsymbol{P}_{f,j}$, is adopted in Eq.~\eqref{eq:enkf}. In this study, $\mathcal{R}=40$ and $\Lambda=1.01$ are applied for $\mathcal{N}=100$, with which the results can always closely follow the truth (see Fig.~\ref{fig:L96state}). For the case of $\mathfrak{N}=10$, we first run a case using the traditional EnKF without inflation and localization, which still fails to follow the truth as shown in Fig.~\ref{fig:L96state}. However, upon integrating the FCNN-based correction term at each analysis step (for which the detailed hyperparameters are shown in Tab.~\ref{tab:parameters}), we observe a remarkable improvement. Fig.~\ref{fig:L96time} compares $\bar{\boldsymbol{s}}^{\mathfrak{N}}_{a}$ of the EnKF-FCNN algorithm against $\bar{\boldsymbol{s}}^{\mathcal{N}}_{a}$, which confirms that the EnKF-FCNN algorithm can still successfully mitigate the filter divergence issue induced by a limited ensemble size for the high-dimensional problem. 

Similar to Lorenz-63, we also evaluate the impact of ensemble size $\mathfrak{N}$, which is varied between $3$ and $80$. Specifically, for each value of $\mathfrak{N}$, we evaluate $\bar{\epsilon}$ given by the EnKF-FCNN algorithm, which is compared to those from the traditional EnKF with and without the optimal set of  inflation factor and localization radius. The results from all three cases are shown Fig.~\ref{fig:epsilon}(b). It can be noted that the EnKF-FCNN algorithm can consistently improve the analysis accuracy for various ensemble sizes without applying the inflation or localization. Moreover, for extremely small $\mathfrak{N}$ (e.g. $\mathfrak{N}=3$), it can still exhibit superior performance compared to the traditional EnKF using $\Lambda_\text{opt}$ and $\mathcal{R}_\text{opt}$ . Afterwards, $\bar{\mathfrak{S}}$ is also quantified for each $\mathfrak{N}$ (Fig.~\ref{fig:mag}(b)), which again shows a monolithic decreasing trend when $\mathfrak{N}$ becomes greater.

In addition, we examine the influence of the observation configurations, including both the observation site number and DA frequency. Specifically, we consider three settings for the available observations, including $\{\chi^{(i)}~|~i=1:1:40\}$, $\{\chi^{(i)}~|~i=1:2:39\}$, and $\{\chi^{(i)}~|~i=1:4:37\}$, and three DA intervals $T_\text{DA}=0.05,~0.10,~\text{and}~0.20~\text{MTU}$. For each configuration, both $\bar{\epsilon}$ and $\bar{\mathfrak{S}}$ are evaluated, for which the results are summarized in Tabs.~\ref{tab:obs96} and~\ref{tab:TDA96}. For all tested cases, the proposed EnKF-FCNN algorithm can improve the overall accuracy, with the error reduced by more than one order of magnitude compared to the traditional EnKF. Also, it can be observed that fewer observations and lower observation frequency will both result in increased $\bar{\epsilon}$, while $\bar{\mathfrak{S}}$ remains relatively constant, which is consistent to the findings for Lorenz-63.  Finally, we evaluate the computational time of running the FCNN function and a single simulation for $T_\text{DA}$ based on Lorenz-96 (Tab.~\ref{tab:CPU-time}), which again confirms that the computational cost induced by FCNN is minimal compared to the forecast step.

\subsection{Nonlinear ocean wave simulation}

\begin{figure}
    \centering
    \begin{subfigure}[h]{0.32\textwidth}
        \centering   \includegraphics[width=\textwidth]{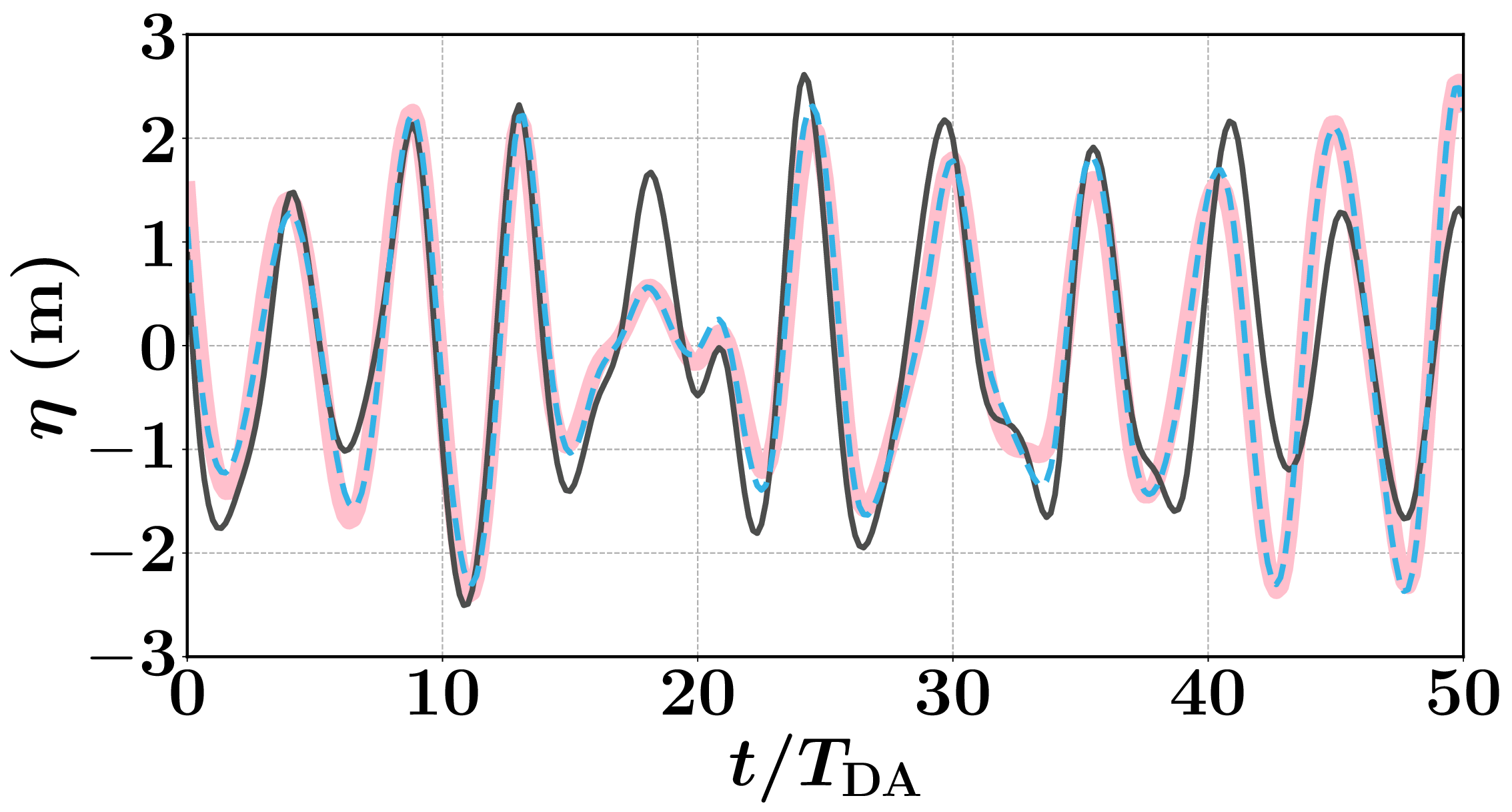}
        \caption{}  
        
    \end{subfigure}  
    \begin{subfigure}[h]{0.32\textwidth}
        \centering     \includegraphics[width=\textwidth]{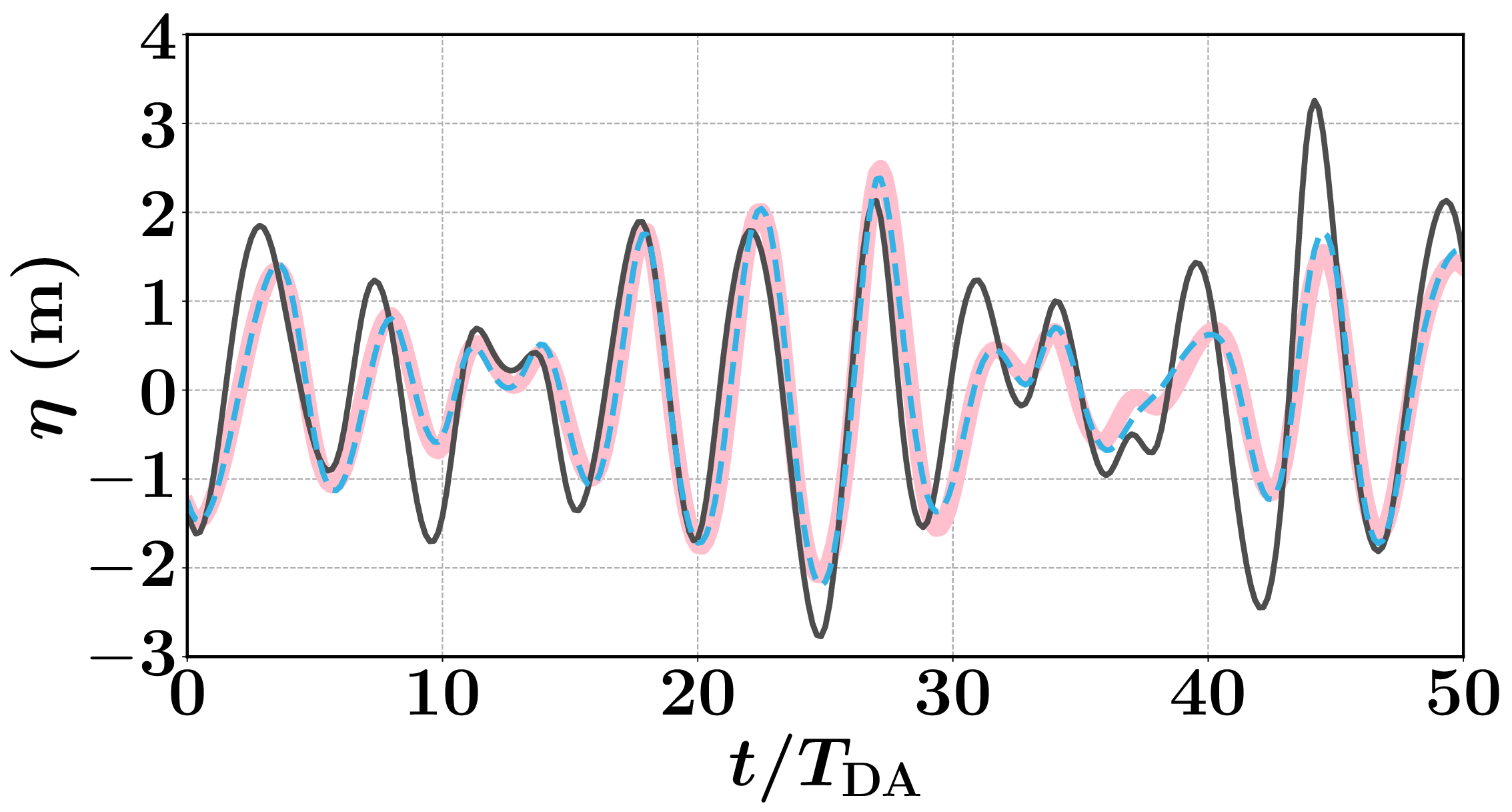}      \caption{}  
        
    \end{subfigure}
    \begin{subfigure}[h]{0.32\textwidth}
        \centering     \includegraphics[width=\textwidth]{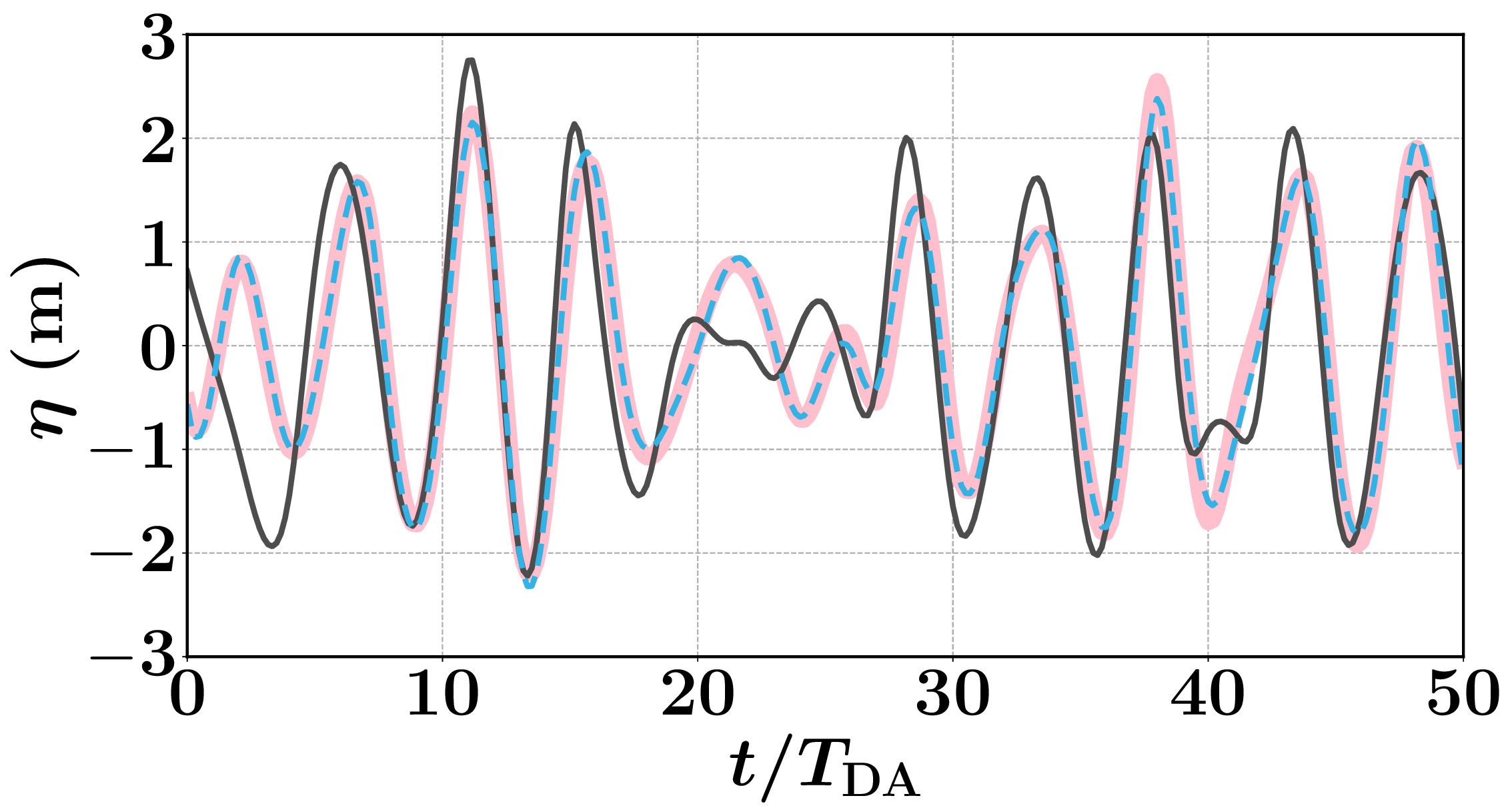}      \caption{}         
    \end{subfigure}
    \caption{Analysis results of $\eta(t)$ obtained with the traditional EnKF using $\mathcal{N}=100$ ({\color{pink}\rule[0.5ex]{0.5cm}{2.0pt}}) and $\mathfrak{N}=20$ ({\color{darkgray}\rule[0.5ex]{0.5cm}{0.5pt}}), as well as the true solution ({\color{cyan}\dashL}), for the 2D wave field: (a)~$x/\mathcal{L}=0.15$, (b)~$x/\mathcal{L}=0.50$, and (c)~$x/\mathcal{L}=0.85$}
     \label{fig:PFL2D-state}
 \end{figure}

\begin{figure}
    \centering
    \begin{subfigure}[h]{0.32\textwidth}
        \centering   \includegraphics[width=\textwidth]{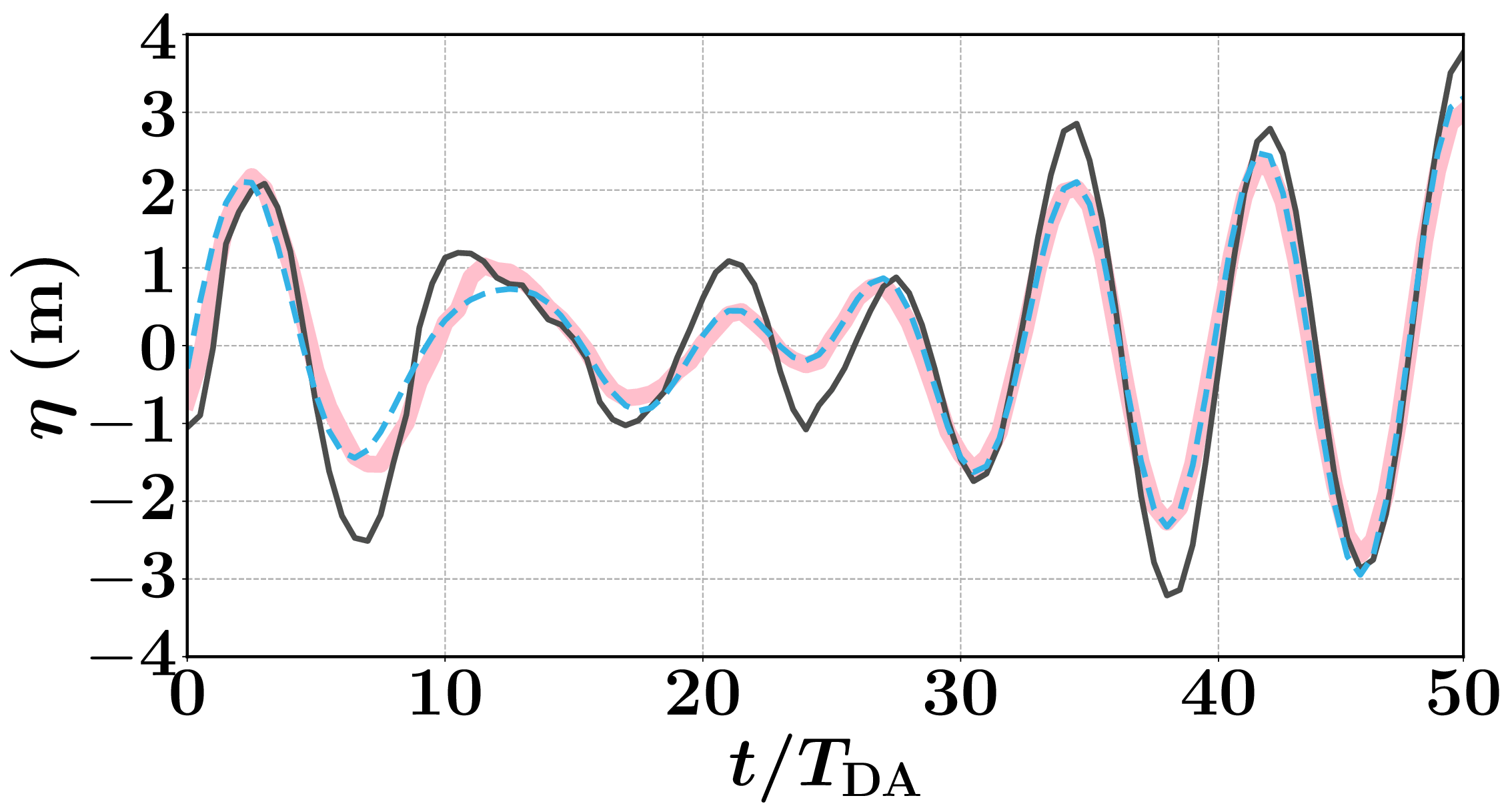}
        \caption{}  
        
    \end{subfigure}  
    \begin{subfigure}[h]{0.32\textwidth}
        \centering     \includegraphics[width=\textwidth]{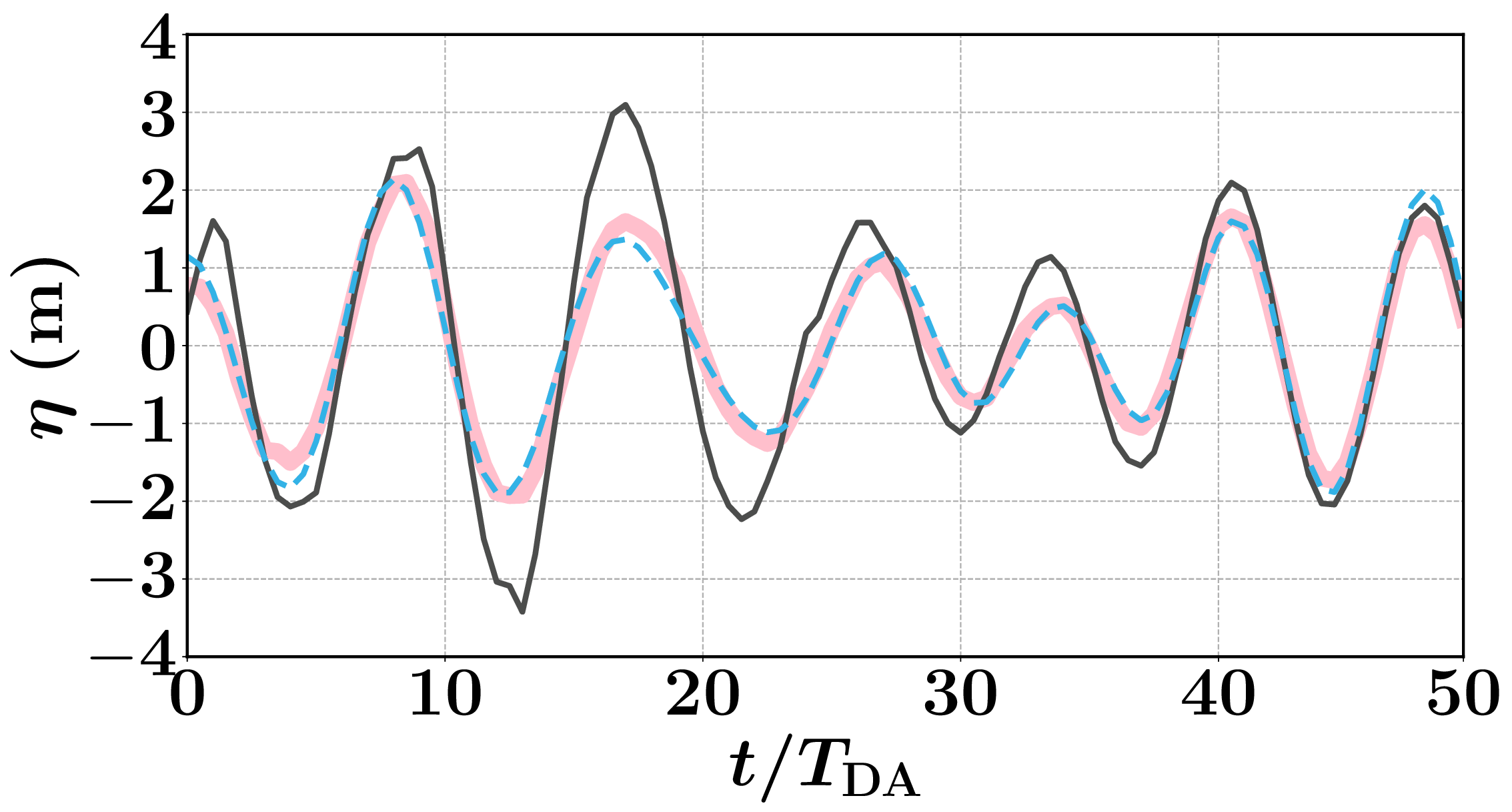}      \caption{}  
        
    \end{subfigure}
    \begin{subfigure}[h]{0.32\textwidth}
        \centering     \includegraphics[width=\textwidth]{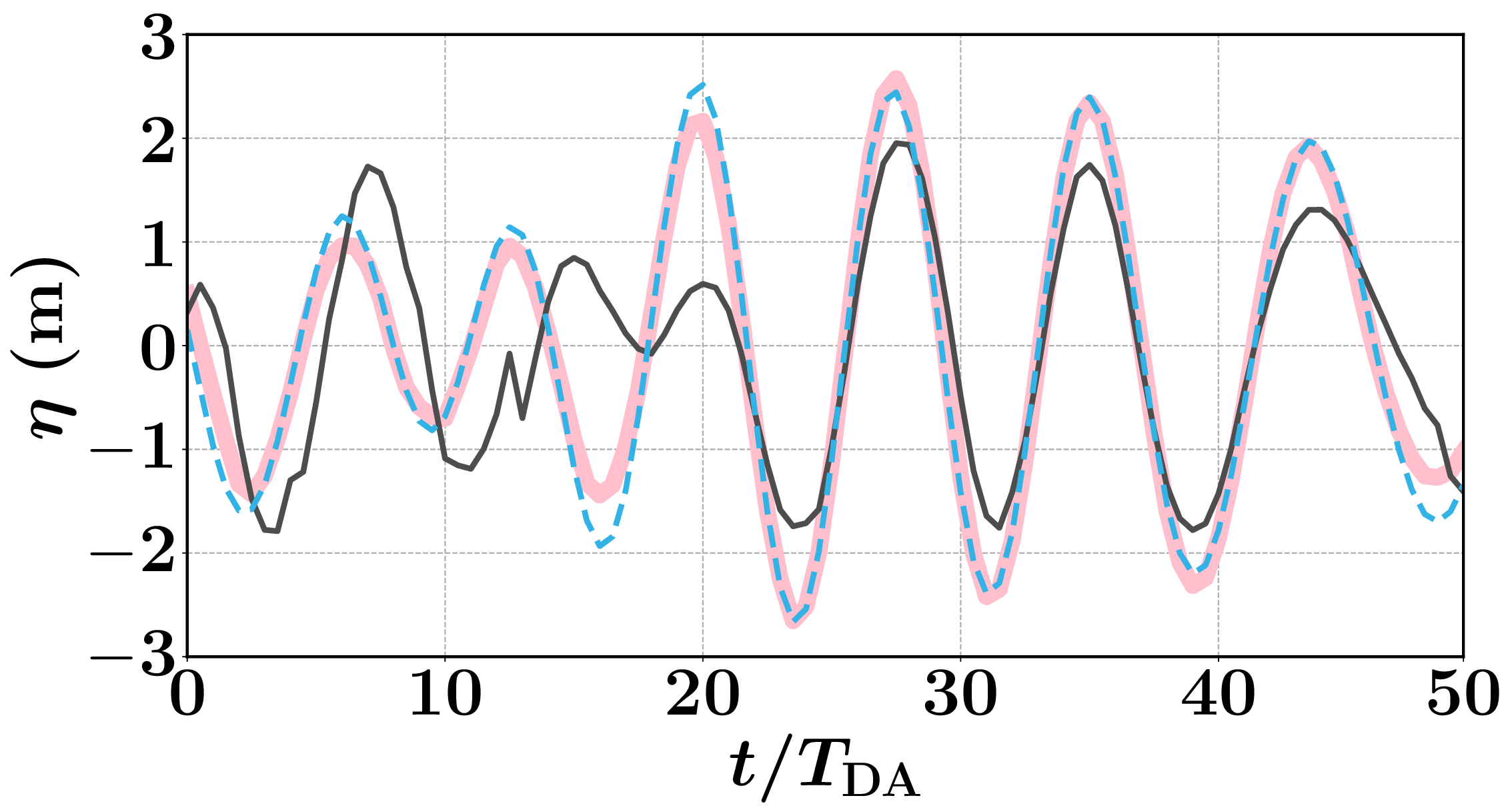}      \caption{}  
        
    \end{subfigure}
    \caption{Analysis results of $\eta(t)$ obtained with the traditional EnKF using $\mathcal{N}=100$ ({\color{pink}\rule[0.5ex]{0.5cm}{2.0pt}}) and $\mathfrak{N}=20$ ({\color{darkgray}\rule[0.5ex]{0.5cm}{0.5pt}}), as well as the true solution ({\color{cyan}\dashL}), for the 3D wave field at $y/\mathcal{L}=0.50$: (a)~$x/\mathcal{L}=0.15$, (b)~$x/\mathcal{L}=0.50$, and (c)~$x/\mathcal{L}=0.85$.}
     \label{fig:PFL3D-state}
 \end{figure}

 \begin{figure}
    \centering
    \begin{subfigure}[h]{0.32\textwidth}
        \centering   \includegraphics[width=\textwidth]{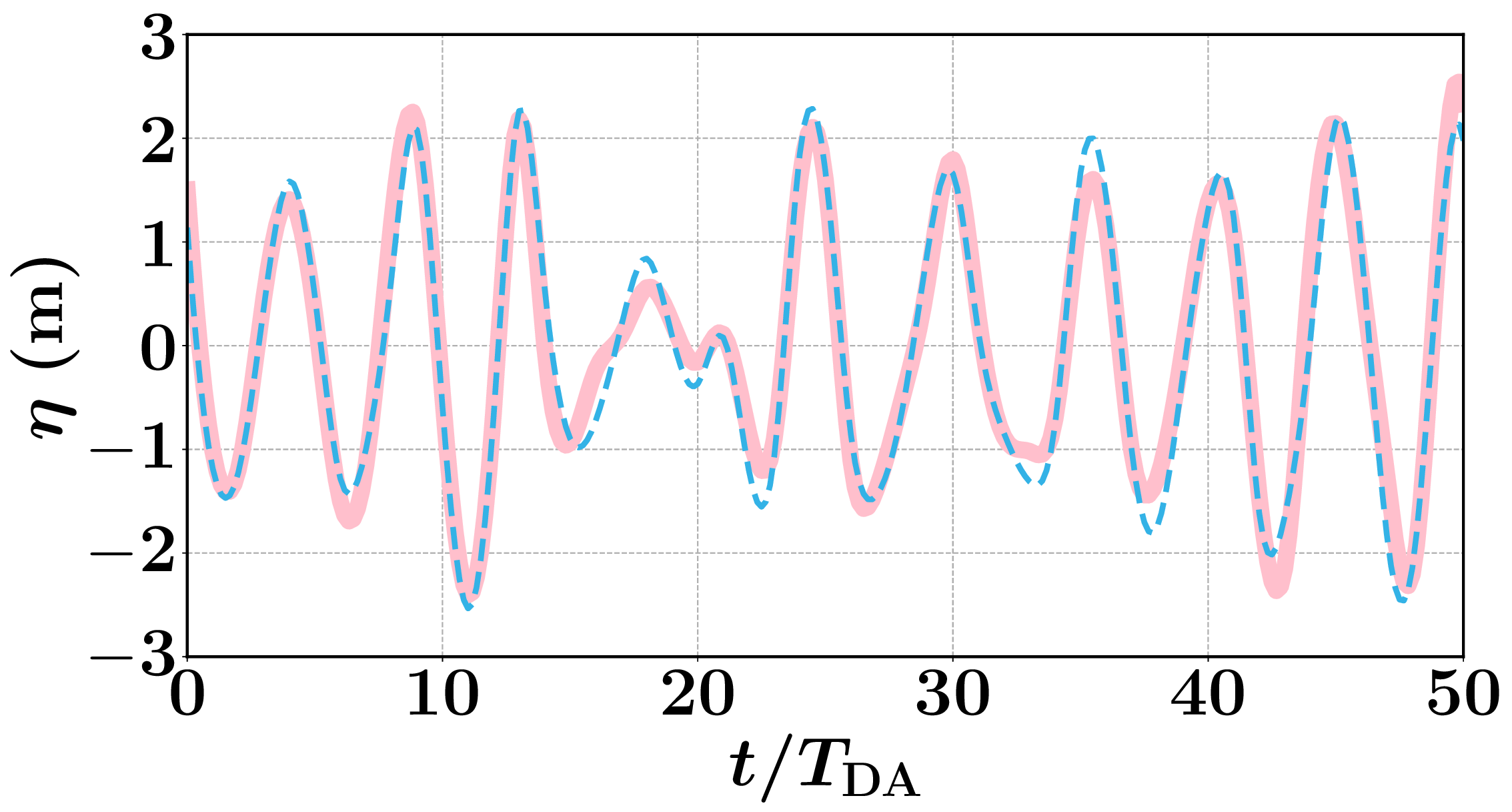}
        \caption{}  
        
    \end{subfigure}  
    \begin{subfigure}[h]{0.32\textwidth}
        \centering     \includegraphics[width=\textwidth]{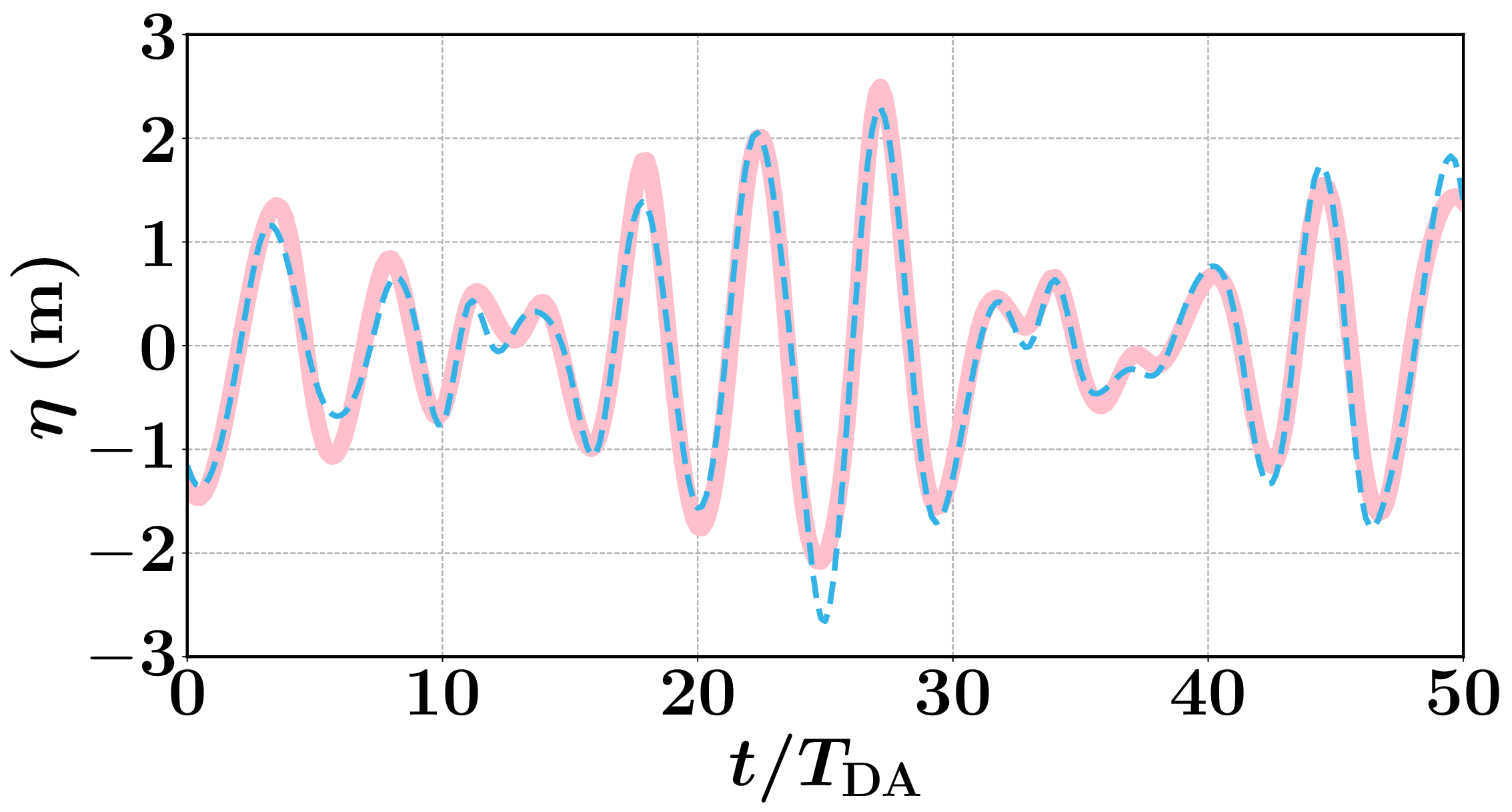}      \caption{}  
        
    \end{subfigure}
    \begin{subfigure}[h]{0.32\textwidth}
        \centering     \includegraphics[width=\textwidth]{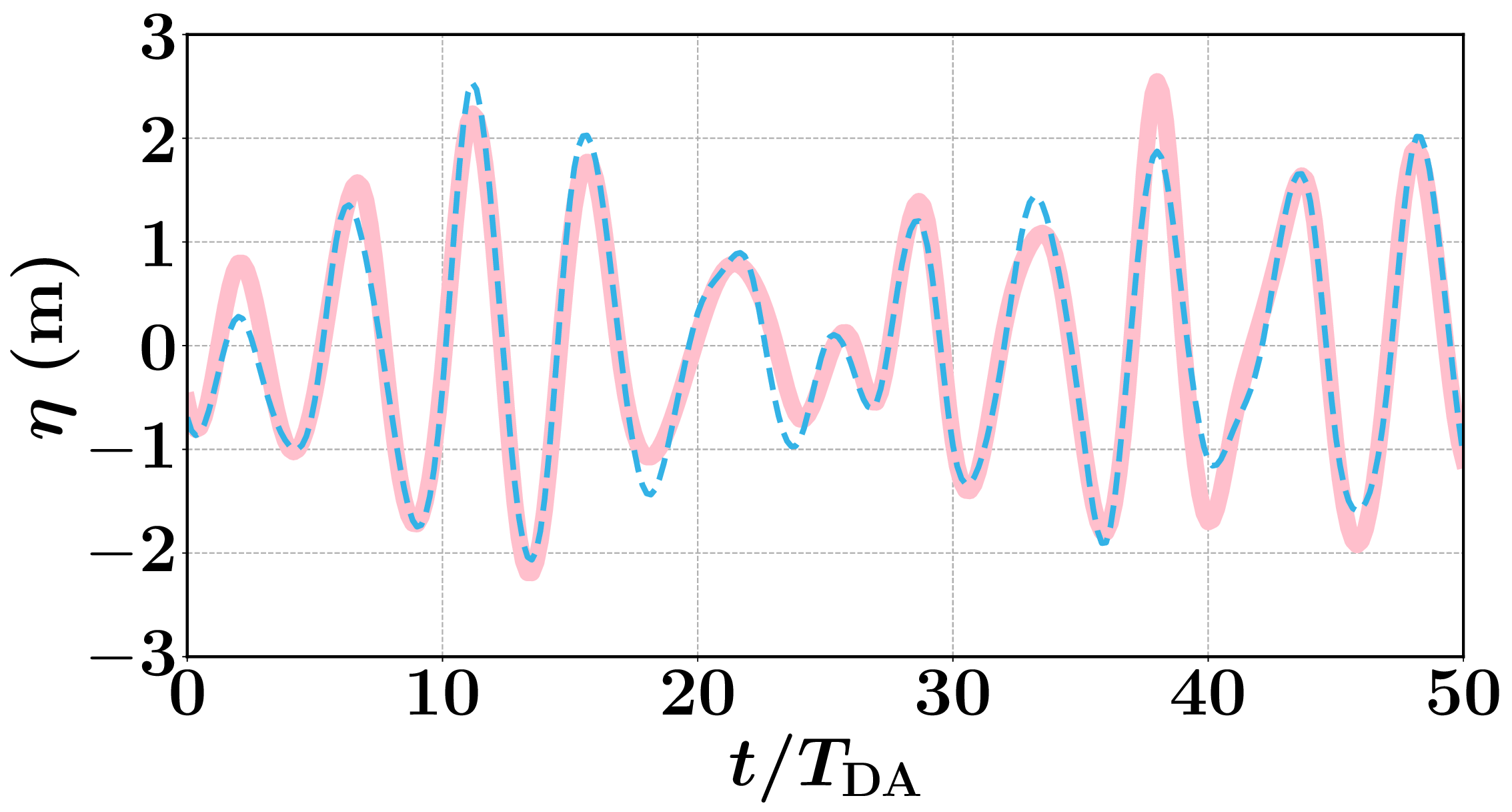}      \caption{}  
        
    \end{subfigure}
    \caption{Analysis results of $\eta(t)$ given by the EnKF-FCNN algorithm with $\mathfrak{N}=20$ ({\color{cyan}\dashL}) and traditional EnKF with $\mathcal{N}=100$ ({\color{pink}\rule[0.5ex]{0.5cm}{2.0pt}}) for the 2D wave field: (a)~$x/\mathcal{L}=0.15$, (b)~$x/\mathcal{L}=0.50$, and (c)~$x/\mathcal{L}=0.85$.}
     \label{fig:PFL2Dtime}
 \end{figure}

 \begin{figure}
    \centering
    \begin{subfigure}[h]{0.32\textwidth}
        \centering   \includegraphics[width=\textwidth]{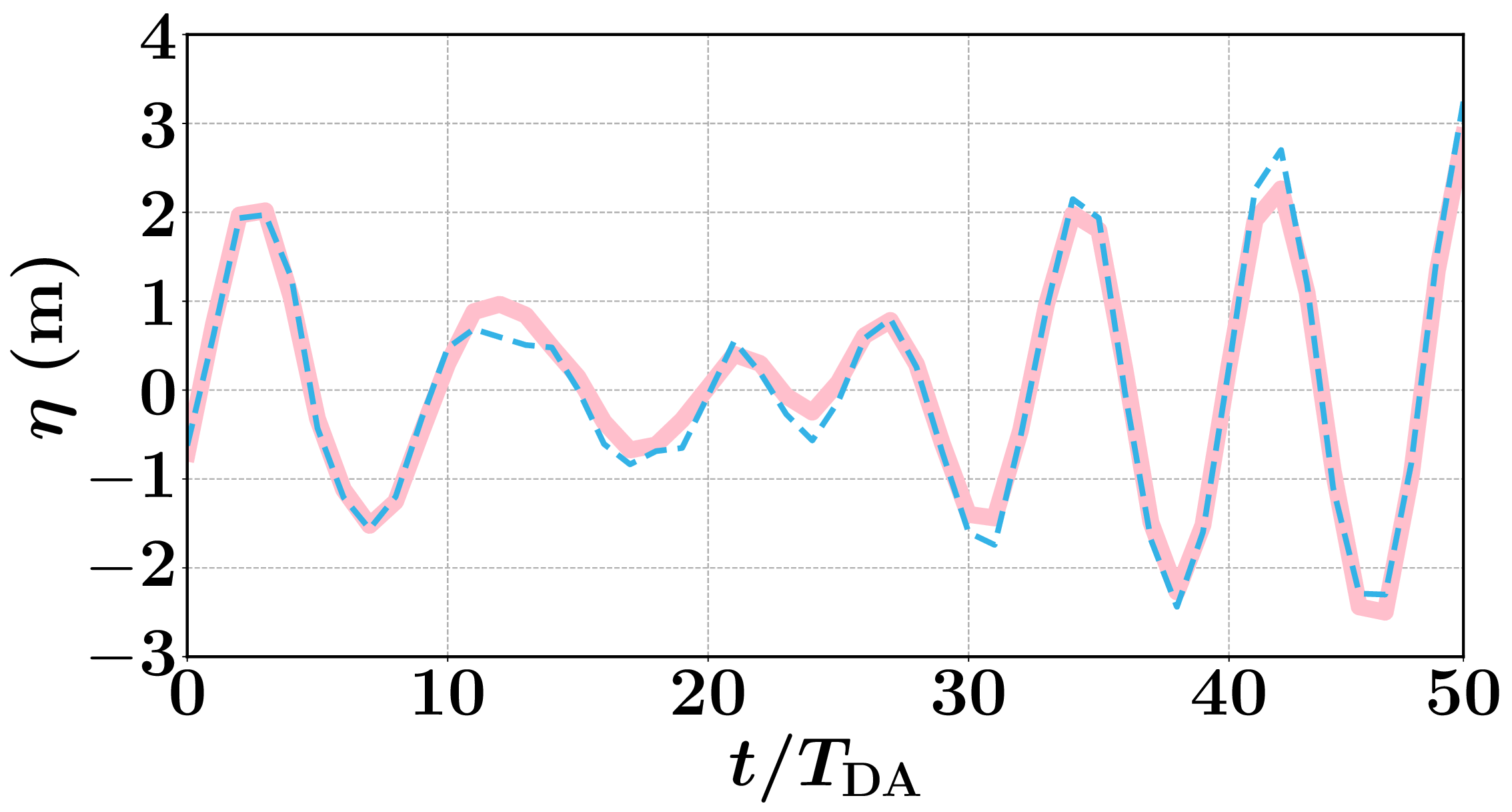}
        \caption{}  
        
    \end{subfigure}  
    \begin{subfigure}[h]{0.32\textwidth}
        \centering     \includegraphics[width=\textwidth]{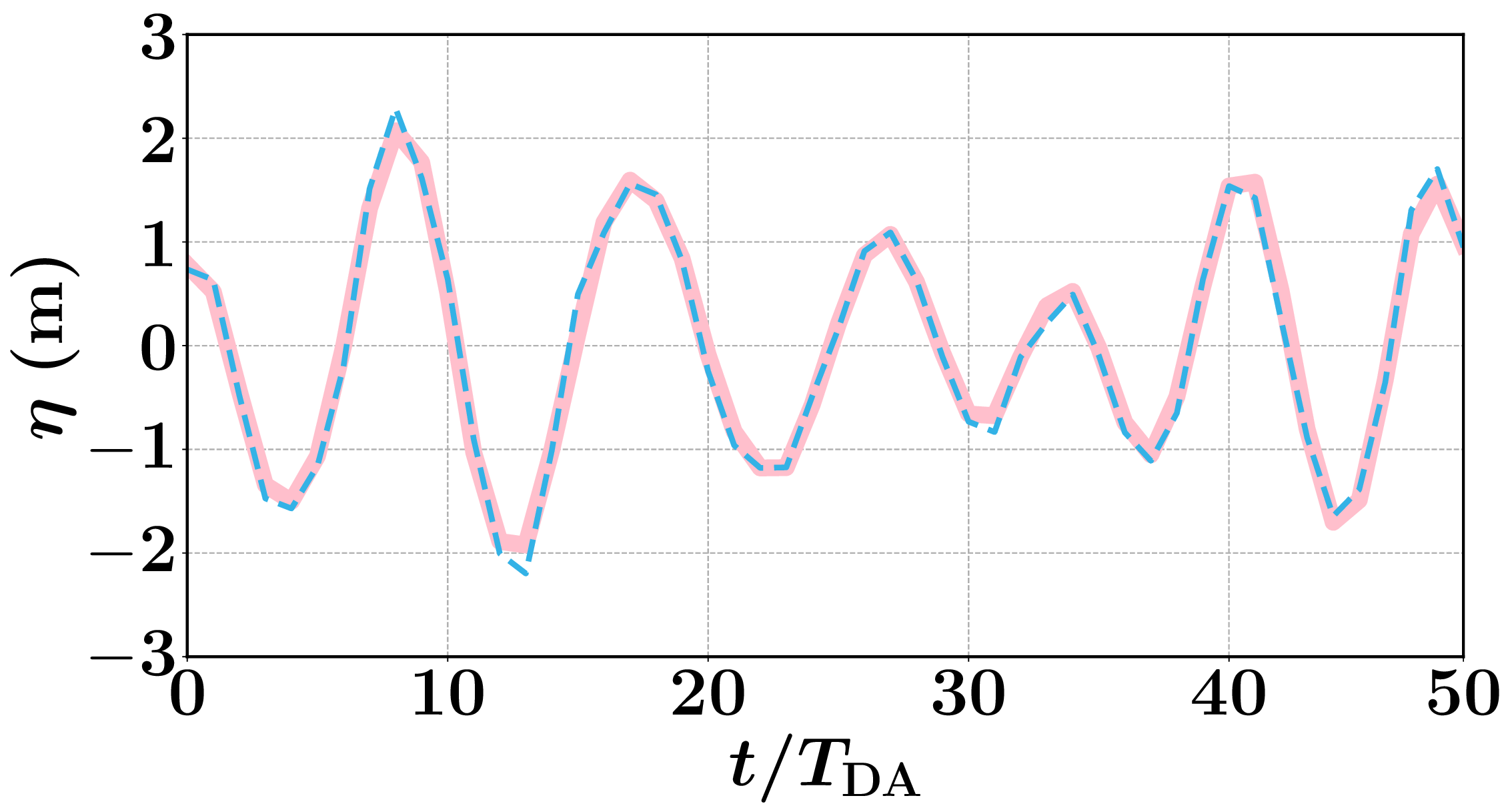}      \caption{}  
        
    \end{subfigure}
    \begin{subfigure}[h]{0.32\textwidth}
        \centering     \includegraphics[width=\textwidth]{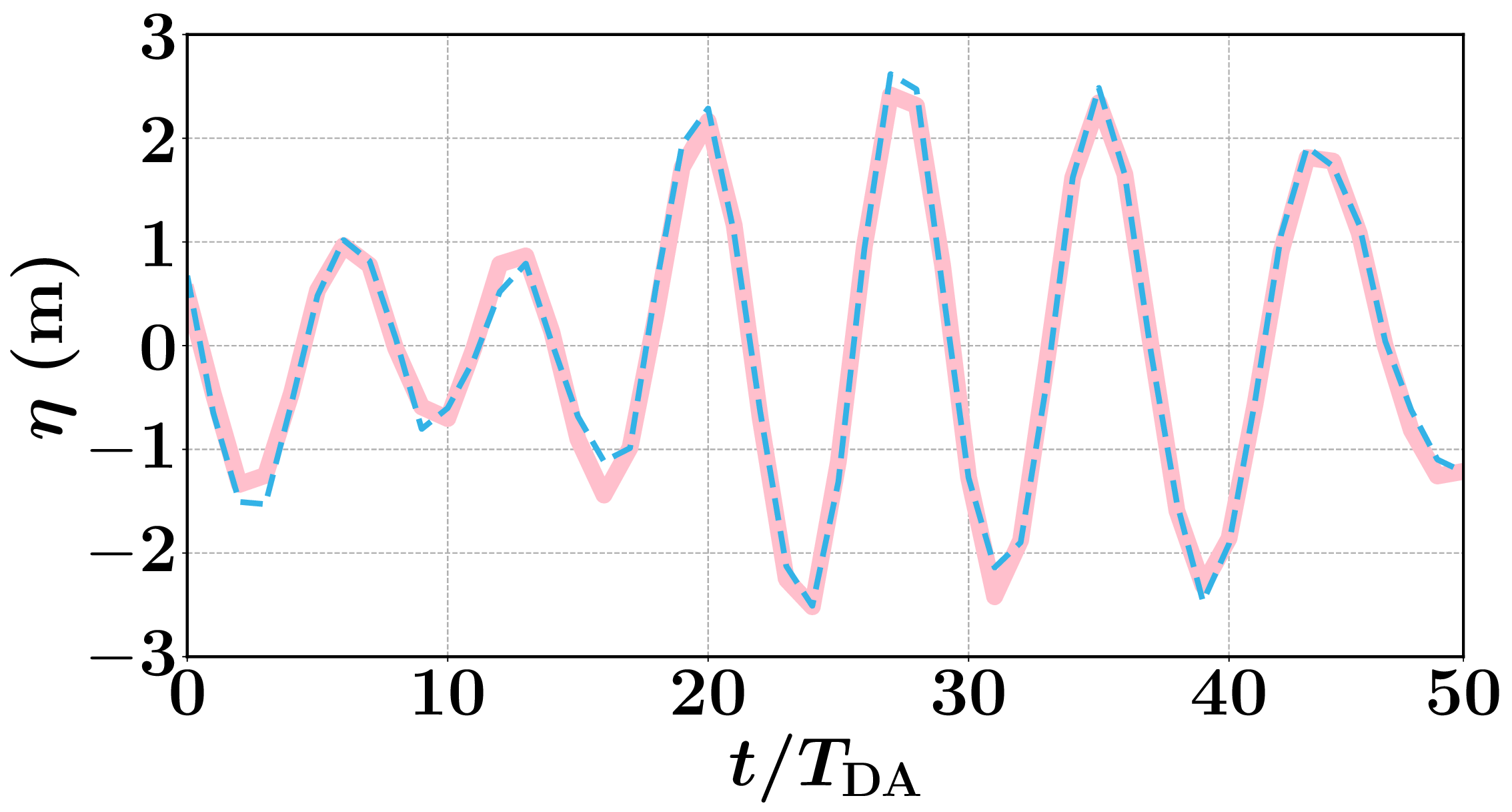}      \caption{}  
        
    \end{subfigure}
    \caption{Analysis results of $\eta(t)$ given by the EnKF-FCNN algorithm with $\mathfrak{N}=20$ ({\color{cyan}\dashL}) and traditional EnKF with $\mathcal{N}=100$ ({\color{pink}\rule[0.5ex]{0.5cm}{2.0pt}}) for the 3D wave field at $y/\mathcal{L}=0.5$: (a)~$x/\mathcal{L}=0.15$, (b)~$x/\mathcal{L}=0.50$, and (c)~$x/\mathcal{L}=0.85$}
     \label{fig:PFL3Dtime}
 \end{figure}

 \begin{figure}
    \centering
    \begin{subfigure}[h]{0.45\textwidth}
        \centering   \includegraphics[width=\textwidth]{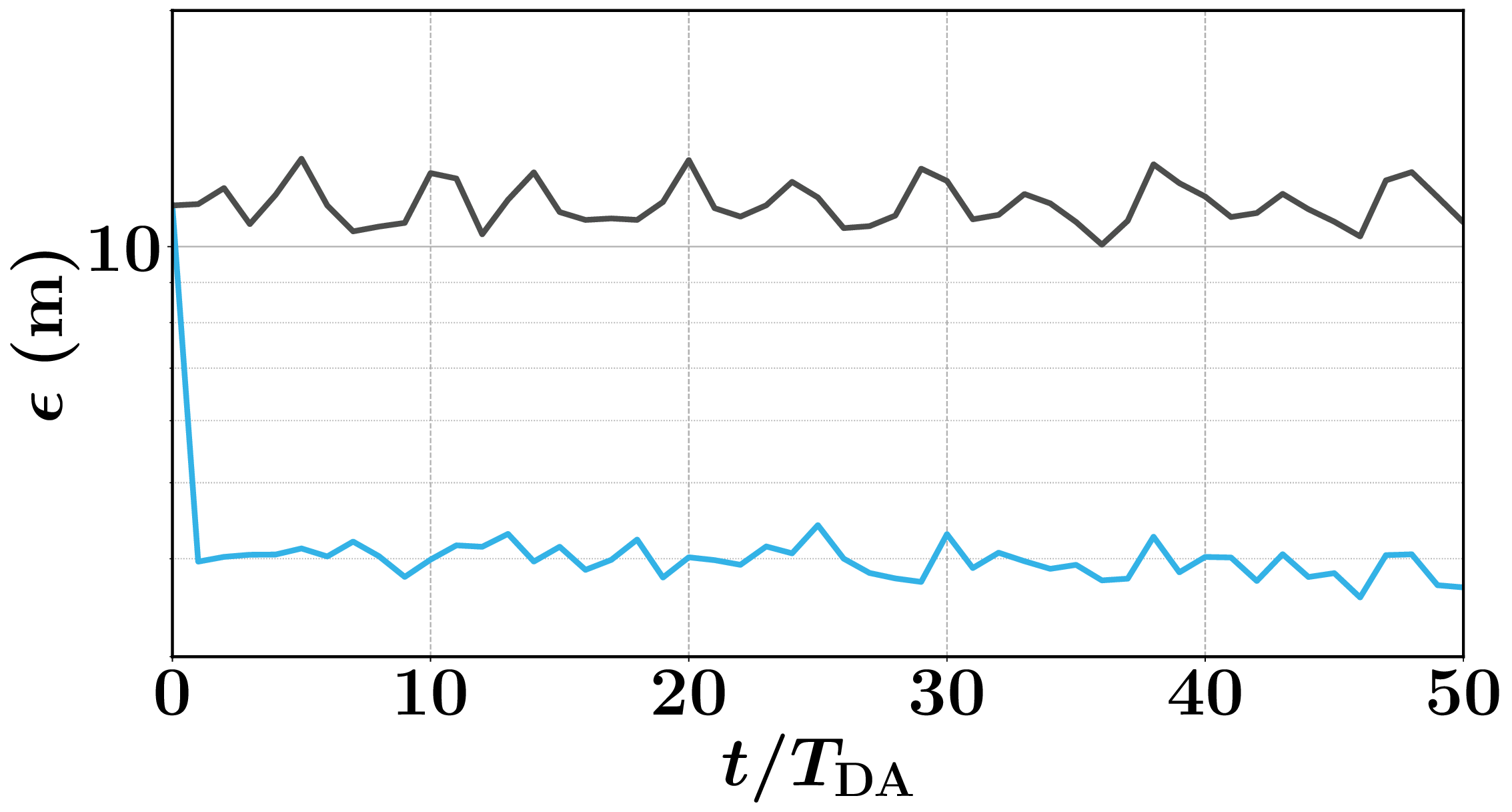}
        \caption{}  
        
    \end{subfigure}  
    \begin{subfigure}[h]{0.45\textwidth}
        \centering     \includegraphics[width=\textwidth]{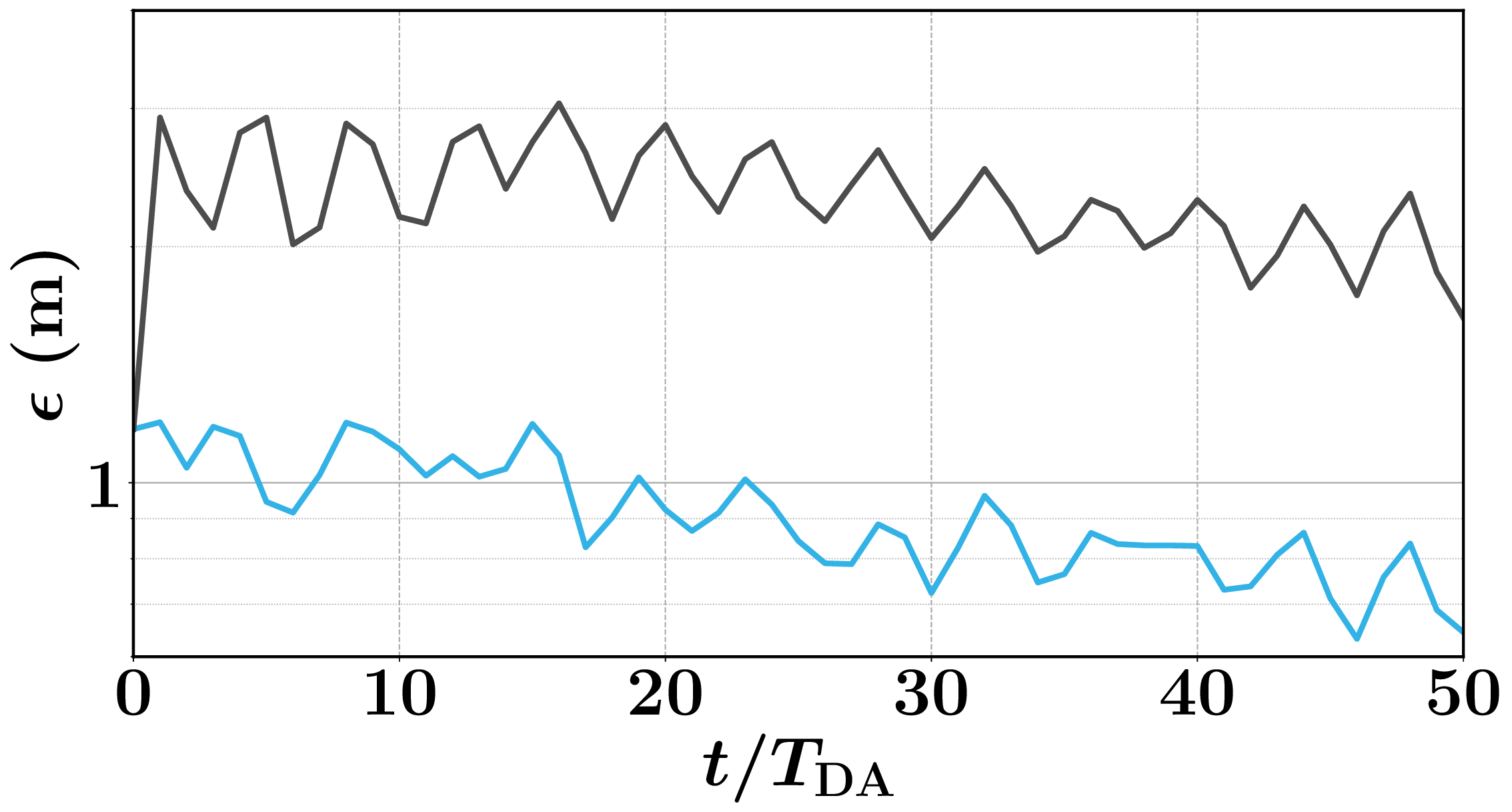}      \caption{}  
        
    \end{subfigure}
    \caption{$\epsilon(t)$ of traditional EnKF ({\color{darkgray}\rule[0.5ex]{0.5cm}{0.5pt}}) and EnKF-FCNN ({\color{cyan}\rule[0.5ex]{0.5cm}{0.5pt}}) with $\mathfrak{N}=20$ for wave fields:: (a)~2D and (b)~3D}
     \label{fig:waveerror}
 \end{figure}

\label{sec:Application}
To further demonstrate the performance of the developed EnKF-FCNN algorithm, we apply it on the nonlinear ocean wave simulation with the PFL method. The governing equations for the ocean wave field are formulated as 
\begin{equation}
\frac{\partial \eta(\boldsymbol{x}, t)}{\partial t} + \frac{\partial \psi(\boldsymbol{x}, t)}{\partial \boldsymbol{x}} \cdot \frac{\partial \eta(\boldsymbol{x}, t)}{\partial \boldsymbol{x}} -
\left[1 + \frac{\partial \eta(\boldsymbol{x}, t)}{\partial \boldsymbol{x}} \cdot \frac{\partial \eta(\boldsymbol{x}, t)}{\partial \boldsymbol{x}}\right] w(\boldsymbol{x}, t) = 0
\label{eq:e1}
\end{equation}
\begin{equation}
\frac{\partial \psi(\boldsymbol{x}, t)}{\partial t} + \frac{1}{2} \frac{\partial \psi(\boldsymbol{x}, t)}{\partial \boldsymbol{x}} \cdot \frac{\partial \psi(\boldsymbol{x}, t)}{\partial \boldsymbol{x}} + g\eta(\boldsymbol{x}, t)-
\frac{1}{2} \left[1 + \frac{\partial \eta(\boldsymbol{x}, t)}{\partial \boldsymbol{x}} \cdot \frac{\partial \eta(\boldsymbol{x}, t)}{\partial \boldsymbol{x}}\right] w(\boldsymbol{x}, t)^2 = 0,
\label{eq:e2}    
\end{equation}
\begin{equation}
\nabla^2 \phi = 0,
\label{eq:e3}
\end{equation}
where $\eta(\boldsymbol{x}, t)$ is the free surface elevation, $\phi$ is the velocity potential, $\psi=\phi|_{z=\eta}$ is the velocity potential on the free surface, $w=(\partial \phi/\partial z)|_{z=\eta}$ is the surface vertical velocity, and $g$ is the gravity acceleration. The key procedure in PFL is to solve Eq.~\eqref{eq:e3} by assuming an artificial boundary condition at one particular depth, with details included in multiple papers such as \citep{klahn2020accuracy,klahn2021simulation}. In particular, we directly adopt the PFL-EnKF framework developed by Wang et al.~\cite{Liu2025} as the basis to test the performance of the EnKF-FCNN coupled algorithm. Both two-dimensional (2D) and three-dimensional (3D) irregular wave fields are considered in this study, with the same case configurations in~\cite{Liu2025}. To produce the dataset, we first randomly generate $100$ initial conditions from a JONSWAP spectrum with the peak steepness $Ak_p=0.12$ and an enhance factor $\gamma=3.3$, and then run the original PFL-EnKF simulations with $\mathcal{N}=100$ and $\mathfrak{N}=20$. Fig.~\ref{fig:PFL2D-state} presents the time histories of surface elevation at three locations ($x/\mathcal{L}=0.15,~0.50,\text{and}~0.85$ with $\mathcal{L}$ being the simulation domain length) of the 2D wave field obtained with the two ensemble sizes, as well as the truth, for one particular initial condition. It can be found that, with $\mathcal{N}=100$, the PFL-EnKF framework can give a good approximation of the truth. In contrast, when using $\mathfrak{N}=20$, the analysis results differ significantly from the truth, especially at the wave crests and troughs. A similar trend can be observed for the 3D wave field, as shown in Fig.~\ref{fig:PFL3D-state}. To build the FCNN, we use the same activation and loss functions as the Lorenz problems, with all other hyperparameters shown in Tab.~\ref{tab:parameters}.

Figures~\ref{fig:PFL2Dtime}-\ref{fig:PFL3Dtime} show the analysis results $\bar{\eta}_a^{\mathfrak{N}}$ given by the EnKF-FCNN algorithm for both 2D and 3D wave fields, respectively, in comparison with $\bar{\eta}_a^{\mathcal{N}}$ produced by the traditional EnKF. It can be found that, for both 2D and 3D wave fields, the EnKF-FCNN algorithm with $\mathfrak{N}=20$ can always demonstrate comparable performance to the traditional EnKF with $\mathcal{N}=100$, which again confirms the EnKF-FCNN's capability of overcoming the filter divergence issue induced by the limited ensemble size. Fig.~\ref{fig:waveerror} compares the results of $\epsilon(t)$ obtained with the traditional EnKF and EnKF-FCNN, in which the ensemble size is always $\mathfrak{N}=20$, for both 2D and 3D wave fields. It can be observed that, by applying the FCNN correction at each analysis step, $\epsilon(t)$ can be consistently reduced by more than $50\%$ for both 2D and 3D wave fields. Similar to the Lorenz systems, we also compare the computational time induced by running the FCNN function with that of running a single simulation for $T_\text{DA}$, with the former approximately three orders of magnitude lower than the latter as shown in Tab.~\ref{tab:CPU-time}, which further confirms the computational efficiency of the proposed EnKF-FCNN algorithm.

\section{Conclusion}
\label{sec:Conclusion}

In this study, we develop a novel EnKF-FCNN coupled algorithm, which aims at improving the robustness of the traditional EnKF method in the scenario of a limited ensemble size. The developed EnKF-FCNN algorithm is validated and tested through a set of numerical experiments based on the numerical simulations of Lorenz systems and nonlinear ocean wave fields. As indicated by the numerical results, the proposed EnKF-FCNN approach enhances the analysis accuracy significantly when working with limited ensemble sizes while introducing minimal additional computational cost. In addition, the developed EnKF-FCNN method can be easily adapted to diverse practical applications, by coupling with different numerical models and substituting the EnKF with other ensemble-based DA methods. 

\section*{Acknowledgements}
This research is financially supported by the National Natural Science Foundation of China (52301336), the Science and Technology Development Fund of Macau S.A.R. (0048/2025/ITP1), the University of Macau (SRG2025-00004-FST), and Zhejiang Provincial Natural Science Foundation of China (LQN25E090004).

\bibliographystyle{elsarticle-harv} 
\bibliography{bibliography}






\end{document}